\documentclass{article}

\PassOptionsToPackage{numbers,sort&compress}{natbib}

\usepackage[main,final]{neurips_2026}
\usepackage{wrapfig}
\usepackage[utf8]{inputenc} 
\usepackage[T1]{fontenc}    
\usepackage{hyperref}       
\usepackage{url}            
\usepackage{booktabs}       
\usepackage{amsfonts}       
\usepackage{nicefrac}       
\usepackage{microtype}      
\usepackage{xcolor}         
\usepackage{subcaption}

\DeclareMathOperator*{\argmax}{arg\,max}

\title{Module-Aware Optimization for Graph Neural Networks}

\author{%
  Guy Hadad \\
  Ben-Gurion University of the Negev \\
  Beer-Sheva, Israel \\
  \texttt{guyhada@post.bgu.ac.il}
  \And
  Haggai Roitman \\
  Ben-Gurion University of the Negev \\
  Beer-Sheva, Israel \\
  \texttt{haggair@gmail.com}
  \AND
  Moshe Eliasof \\
  Ben-Gurion University of the Negev \\
  Beer-Sheva, Israel \\
  \texttt{eliasof@bgu.ac.il}
  \And
  Bracha Shapira \\
  Ben-Gurion University of the Negev \\
  Beer-Sheva, Israel \\
  \texttt{bshapira@bgu.ac.il}
}

\begin{document}

\maketitle

\begin{abstract}
Different optimizers impose different geometries on parameter updates. Adaptive methods such as
Adam and AdamW use coordinate-wise updates, while matrix-aware optimizers such
as Muon utilize weight-matrix structure through orthogonalized updates. For
Graph Neural Networks (GNNs), however, the effective geometry of its weight matrices is
shaped not only by its learned channel-mixing weights, but also by the graph-propagated
representations it acts on. Thus, even if an update is well normalized as a matrix, its effect on the
layer output can be amplified when applied to graph-propagated features. To address this gap, we introduce {GraphMuon}, a graph-aware optimizer
that adapts Muon's matrix updates to the covariance structure of
graph-propagated layer inputs. GraphMuon performs orthogonalization in a metric
defined by these propagated representations and normalizes step sizes according
to the update's induced effect on the layer representation. We show that 
theoretically, vanilla Muon can suffer from graph-induced update amplification,
while our GraphMuon can mitigate this effect. As such, our GraphMuon yields a graph-metric analogue of
Muon's polar update step. Empirically, GraphMuon achieves
$1.32\times$--$2.42\times$ validation-loss time-to-target
speedups over AdamW on three large-scale OGB benchmarks.
Across 15 dataset--architecture configurations spanning
node classification, inductive graph classification, and
graph regression, it achieves lower time-to-target in
13 cases, while maintaining competitive or superior
predictive performance. Code is available at \url{https://github.com/guy-hadad/GraphMuon}.
\end{abstract}



\section{Introduction}
\label{sec:introduction}

Optimizers define the geometry in which neural-network updates are measured.
Classical SGD uses a global Euclidean scale \citep{robbins1951stochastic},
adaptive methods such as Adam and AdamW normalize updates coordinate-wise
\citep{kingma2014adam,loshchilov2017decoupled}, and preconditioned methods such
as Shampoo exploit matrix statistics to approximate richer curvature information
\citep{gupta2018shampoo}. More recently, matrix-aware optimizers such as Muon
treat weight matrices as structured operators and replace raw momentum updates
with approximately orthogonalized matrix directions \citep{jordan2024muon}.
Muon has also been shown to improve compute efficiency in large language-model
training \citep{liu2025muon,team2026kimi,team2025kimi}. These developments
suggest a broader principle: the geometry used by an optimizer should depend not
only on the shape of a parameter tensor, but also on the module in which that
parameter appears \citep{bernstein2025manifolds}.

Graph neural networks provide a setting where this module dependence is
especially important. In a standard feed-forward layer, a weight matrix acts on
representations produced by previous neural layers. In a GNN layer, however, the
trainable matrix acts on graph-propagated representations. For a propagation
operator $S$ and node representation $H$, the effective input to the weight
matrix is
\begin{equation}
    X_g = SH.
\end{equation}
Therefore, the effect of an update $\Delta W$ is governed by the induced
representation-level perturbation $X_g\Delta W$, rather than by the
parameter-space norm of $\Delta W$ alone. Since $X_g$ depends on the graph
topology, neighborhood aggregation, feature smoothing, and the current training
state, two updates with the same Euclidean matrix scale can induce very different
changes in the layer output.

This observation exposes a gap in existing optimizers for GNNs. AdamW adapts
updates through coordinate-wise gradient moments, but does not explicitly
distinguish directions that are amplified by graph propagation from directions
that are attenuated in the propagated feature distribution. Muon improves over
coordinate-wise updates by using the matrix structure of $W$, but its
orthogonalization is still input-agnostic: the same Euclidean matrix geometry is
used whether the matrix acts on dense features, tokens, or graph-propagated node
representations. For GNNs, the relevant object is not only the standalone matrix
$W$, but the graph-conditioned map
\begin{equation}
    W \mapsto X_g W.
\end{equation}

The geometry of this map is determined by the covariance of the propagated
inputs,
\begin{equation}
    C_g = \frac{1}{n}X_g^\top X_g.
\end{equation}
Directions aligned with high-energy eigenspaces of $C_g$ produce large
representation-level perturbations, while directions in low-energy eigenspaces
have much smaller immediate effect \citep{horn2012matrix}. Thus, an optimizer for GNN weight matrices
should be both matrix-aware and graph-aware: it should exploit the matrix
structure of $W$, while calibrating update direction and scale to the
graph-induced geometry of the representations on which $W$ actually acts.

Inspired by this observation, we introduce \textbf{\method{}}, a graph-aware extension of Muon for GNN
optimization. \method{} adapts Muon's matrix-orthogonalized updates to the
layer-wise geometry induced by graph-propagated inputs. It has two complementary
components. First, \emph{graph-aware orthogonalization} applies the polar-style
Muon update in a metric defined by the covariance of $X_g$. Second,
\emph{graph-aware step normalization} scales each update by its induced
representation-level perturbation, rather than by its parameter-space magnitude
alone. Together, these mechanisms align the optimizer geometry with the actual
forward map of a GNN layer.

Our analysis in Section \ref{subsec:theory} shows that this modification is principled: a vanilla Muon update,
although normalized in Euclidean matrix space, can still be amplified in
representation space by the dominant eigenspaces of $C_g$. In contrast, our
\method{} directly controls the quantity
$\|X_g\Delta W\|_F/\sqrt{n}$, which is the perturbation induced in the
pre-activation representation of the GNN layer. We further show that the
graph-aware orthogonalization step is the graph-metric analogue of Muon's polar
update, rather than an ad hoc rescaling.

\begin{figure}[t]
    \centering
    \includegraphics[width=0.9\linewidth]{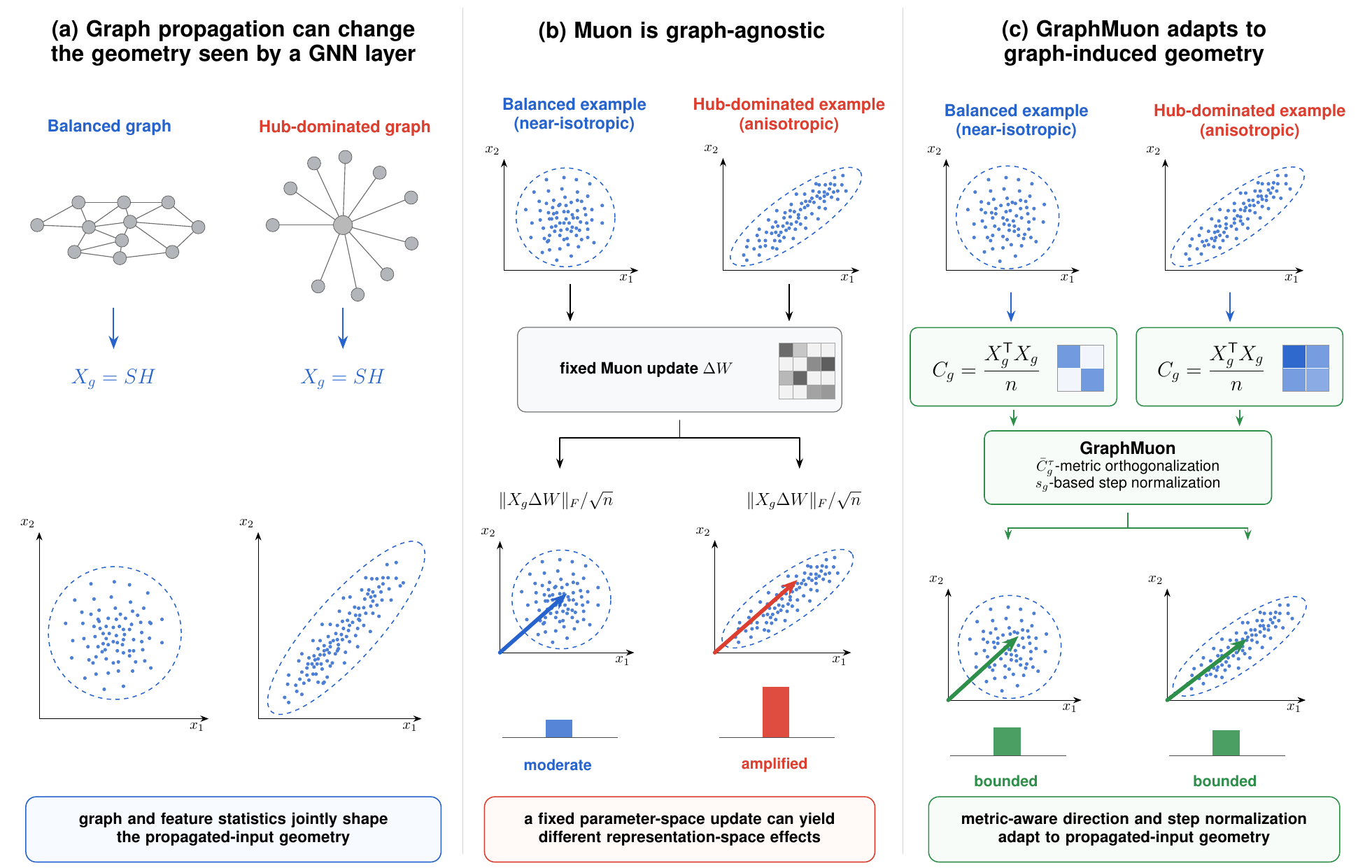}
    \caption{
    Motivation for graph-aware optimization.
    A GNN weight matrix acts on the graph-propagated input $X_g = SH$, whose
    geometry depends on graph structure and propagated feature statistics.
    As a result, a Muon update that is well normalized in Euclidean matrix space
    can still produce an amplified representation-level perturbation
    $\|X_g \Delta W\|_F$.
    \method{} addresses this mismatch by adapting the update geometry to the
    graph-induced covariance $C_g = X_g^\top X_g / n$ and scaling each step by
    its induced effect on graph-propagated representations.
    }
    \label{fig:graphmuon_motivation}
\end{figure}

Figure~\ref{fig:graphmuon_motivation} illustrates the resulting mismatch and
our correction. A Muon update can be well normalized as a matrix while still
causing a large graph-propagated perturbation. \method{} addresses this by using
$C_g$ to adapt both the update geometry and its scale.

We evaluate GraphMuon across graph-learning benchmarks
spanning node classification, inductive graph classification,
and graph regression, using GCN, GAT, GINE, and
GraphGPS architectures. Under a fixed validation-loss
time-to-target protocol, GraphMuon achieves
$1.32\times$--$2.42\times$ wall-clock speedups over AdamW
on the three large-scale OGB benchmarks. Across a broader
evaluation of 15 dataset--architecture configurations,
GraphMuon improves epoch-level time-to-target in all cases
and achieves faster wall-clock convergence in 13 of 15,
while maintaining competitive or superior predictive
performance.

\textbf{Main Contributions.}
Our contributions are as follows:
\begin{enumerate}
  \item We identify a \emph{graph-induced optimization mismatch}: standard
  optimizers normalize updates in parameter space, whereas the effect of a GNN
  weight update is determined by its perturbation of graph-propagated
  representations.

  \item We introduce \method{}, a graph-aware extension of Muon that adapts
  matrix-update geometry to the layer-wise covariance of graph-propagated
  representations and normalizes updates by their representation-level effect.

  \item We prove that \method{} controls graph-induced representation
  perturbations and yields a graph-metric analogue of Muon's polar update.

    \item We demonstrate faster validation-loss convergence
    across diverse graph-learning regimes, including
    inductive graph-level tasks and multiple GNN architectures,
    with wall-clock improvements in 13 of 15 evaluated
    configurations and competitive or superior predictive
    performance.

\end{enumerate}

\section{Related Work}
\label{sec:related_work}

\textbf{Graph neural networks.}
GNNs extend deep learning to graph-structured data by interleaving learnable
transformations with neighborhood aggregation~\citep{scarselli2008graph,gilmer2017neural}.
Scalable spatial methods, including GCN~\citep{kipf2016semi},
GraphSAGE~\citep{hamilton2017inductive}, GAT~\citep{velivckovic2018graph},
and GIN~\citep{xu2018powerful}, have become the dominant paradigm, with
subsequent work extending these ideas to heterogeneous graphs~\citep{schlichtkrull2018modeling}
and heterophilous graphs where connected nodes tend to differ in
class~\citep{zhu2020beyond}. Across these architectures, weight matrices act
not on raw node features but on representations smoothed and mixed by graph
propagation. This distinguishes GNN training from standard deep-network
optimization: the geometry seen by each weight matrix is shaped not only by
network depth, but also by graph topology and its induced propagation
operator, which is precisely the mismatch \method{} is designed to address.

\textbf{Matrix-aware optimization.}
Classical SGD and momentum update parameters in Euclidean space, while
AdaGrad, Adam, and AdamW normalize updates coordinate-wise using accumulated
gradient statistics
\citep{duchi2011adaptive,kingma2014adam,loshchilov2017decoupled}. Structured
optimizers exploit richer geometry: K-FAC uses Kronecker-factored curvature
\citep{martens2015optimizing}, Shampoo uses matrix preconditioners
\citep{gupta2018shampoo}, and Muon applies momentum followed by approximate
matrix orthogonalization~\citep{jordan2024muon}. Muon is the closest
predecessor, but remains module-agnostic: it uses the same Euclidean matrix
geometry for dense features, tokens, and graph-propagated representations.
\method{} instead tests whether update geometry should match the
module-induced operator, which for GNNs is the graph-conditioned map
\(W \mapsto X_g W\), rather than the standalone matrix \(W\).

\textbf{Graph-aware modules and optimization.}
A central theme in GNN research is adapting neural-network components to graph
structure: GCNs~\citep{kipf2016semi}, GAT~\citep{velivckovic2018graph}, and
GIN~\citep{xu2018powerful} do so architecturally, as do graph-specific
normalization and modular designs~\citep{cai2021graphnorm,rampavsek2022recipe}.
Izadi et al.~\citep{izadi2020optimization} show that optimization itself can
be graph-aware by deriving natural-gradient updates for graph-based
semi-supervised learning. \method{} follows this optimizer-level view but
targets a different setting: rather than a natural-gradient method, it adapts
Muon's matrix-aware geometry to the layer-wise covariance of graph-propagated
representations, the optimizer-level analogue of graph-adaptive GNN design.
Additional related work is discussed in Appendix~\ref{app:additional_related_work}.

\section{Background and Preliminaries}
\label{sec:background}

\textbf{Notations.}
Let $G=(V,E)$ be a graph with adjacency matrix $A\in\mathbb{R}^{n\times n}$ and
node representations $H^{(\ell)}\in\mathbb{R}^{n\times d_\ell}$. Adding
self-loops and symmetrically normalizing gives:
\[
    S = \widetilde{D}^{-1/2}\widetilde{A}\,\widetilde{D}^{-1/2},
    \qquad \widetilde{A}=A+I,
\]
where $\widetilde{D}$ is the degree matrix of $\widetilde{A}$. Each GNN weight
matrix $W^{(\ell)}$ acts on the \emph{graph-propagated} representation
$X_g^{(\ell)} = SH^{(\ell)}$, so the geometry seen by $W^{(\ell)}$ is shaped by
both the node features and the graph structure in $S$.

\textbf{Muon.}
Muon~\citep{jordan2024muon} is an optimizer for matrix-valued parameters. Given
gradient $G_t = \nabla_W \mathcal{L}(W_t)$, it maintains momentum
$B_t = \mu B_{t-1} + G_t$ and forms $M_t = G_t + \mu B_t$ (Nesterov) or
$M_t = B_t$ otherwise. It replaces $M_t$ with an approximately orthogonalized
direction via $K$ Newton--Schulz (NS) iterations,
\begin{equation}
    D_t = \mathrm{NS}_K(M_t),
    \label{eq:muon_dir}
\end{equation}
and updates $W_{t+1} = W_t - \eta D_t$ (optionally with decoupled weight decay;
see Appendix~\ref{app:muon_details}). The key property is that $D_t$ has more
balanced singular values than $M_t$. However, vanilla Muon treats $W$ as a
generic matrix and does not account for the graph-induced geometry of
$X_g^{(\ell)}$. For additional background, see Appendix~\ref{app:full_background}.

\section{\method}
\label{sec:method}
In this section we formalize \method{} - a Muon-based optimizer specialized for GNNs. Because $W^{(\ell)}$ acts on graph-propagated features, i.e., $X_g^{(\ell)} = SH^{(\ell)}$,
the impact of an update depends on the graph-induced feature geometry. We adapt
Muon in two ways: \emph{graph-aware orthogonalization} (Section~\ref{sec:graph_aware_orthogonalization}) adjusts the update
direction using this geometry, and \emph{graph-aware step normalization} (Section~\ref{sec:graph_aware_step_normalization}) controls
the resulting representation-level perturbation. We conclude this section with a discussion on \method{}'s theoretical properties (Section~\ref{subsec:theory}). 

\subsection{Graph-aware Orthogonalization}
\label{sec:graph_aware_orthogonalization}

For layer $\ell$, define the graph-induced covariance
$C_g^{(\ell)} = \tfrac{1}{n}X_g^{(\ell)\top}X_g^{(\ell)}$,
where $1/n$ removes dependence on graph size. Since $C_g^{(\ell)}$ may be
singular or ill-conditioned, we use the trace-normalized and regularized matrix:
\begin{equation}
    \bar{C}_g^{(\ell)}
    =
    \frac{C_g^{(\ell)}}{\operatorname{tr}(C_g^{(\ell)})/d_\ell}
    + \epsilon_{\mathrm{cov}} I,
\end{equation}
where $d_\ell$ is the input dimension of $W^{(\ell)}$ and
$\epsilon_{\mathrm{cov}}>0$ is a small regularizer. We compute
$(\bar{C}_g^{(\ell)})^{\tau/2}$ via eigendecomposition
$\bar{C}_g^{(\ell)} = V\Lambda V^\top$, caching eigenbases and recomputing only
when $\|\Delta\bar{C}_g^{(\ell)}\|_F > \delta$. For $d_\ell > d_{\max}$, we
substitute the diagonal approximation $\operatorname{diag}(\bar{c}_i^{\tau/2})$,
reducing cost from $O(d_\ell^3)$ to $O(d_\ell)$.

Graph-aware orthogonalization maps the momentum into the graph metric,
orthogonalizes, then maps back:
\vspace{-0.1in}
\begin{equation}
    D_{g,\tau,t}^{(\ell)}
    =
    \left(\bar{C}_g^{(\ell)}\right)^{-\tau/2}
    \mathrm{NS}_K\!\left(
    \left(\bar{C}_g^{(\ell)}\right)^{\tau/2} M_t^{(\ell)}
    \right).
    \label{eq:graph_aware_orthogonalization}
\end{equation}
The hyperparameter $\tau\in[0,1]$ controls graph-awareness strength. In particular, $\tau=0$
implements standard Muon, while larger $\tau$ increasingly respects the
graph-propagated feature geometry. The sensitivity of \method{} to this hyperparameter is studied in Appendix~\ref{app:tau_interpolation}.

\subsection{Graph-aware Step Normalization}
\label{sec:graph_aware_step_normalization}

Because $W^{(\ell)}$ acts on $X_g^{(\ell)}$, update size should be measured by the
induced representation perturbation $\|X_g^{(\ell)}\Delta W^{(\ell)}\|_F$, not
by $\|\Delta W^{(\ell)}\|_F$ alone. We use the \emph{raw} (not trace-normalized)
covariance $C_g^{(\ell)}$ to preserve the true layer-wise feature scale (see
Appendix~\ref{app:step_norm_derivation}). Given update direction $D^{(\ell)}$,
GraphMuon computes the graph-aware scale:
\begin{equation}
    s_g^{(\ell)}
    =
    \sqrt{
        \frac{
            \operatorname{tr}\!\left(D^{(\ell)\top} C_g^{(\ell)} D^{(\ell)}\right)
        }{
            \|D^{(\ell)}\|_F^2
        }
        + \epsilon_{\mathrm{scale}}
    },
    \label{eq:graph_scale}
\end{equation}
where $\epsilon_{\mathrm{scale}}>0$ is a small constant. Since
$\operatorname{tr}(D^{(\ell)\top}C_g^{(\ell)}D^{(\ell)}) =
\tfrac{1}{n}\|X_g^{(\ell)}D^{(\ell)}\|_F^2$, the scale $s_g^{(\ell)}$ can be
computed directly from graph-propagated features without forming $C_g^{(\ell)}$.

Following the original Muon implementation, we apply the shape-dependent
correction $r = \max(1,\, d_\ell/d_{\ell+1})^{1/2}$, which compensates for the
scale change introduced by a semi-orthogonal update on rectangular matrices.
The  GraphMuon update is given by:
\begin{equation}
    W_{t+1}^{(\ell)}
    =
    \bigl(1 - \eta\,r\lambda\bigr)W_t^{(\ell)}
    -
    \eta\,r\,\frac{D^{(\ell)}}{s_g^{(\ell)}},
    \label{eq:graphmuon_update}
\end{equation}
where $\eta$ is the learning rate and $\lambda$ is the decoupled weight-decay
coefficient.

\subsection{Theoretical Properties}
\label{subsec:theory}


We summarize the main theoretical properties of \method{}: it controls the
perturbation induced in the representation space actually used by each GNN
layer, and its graph-aware orthogonalization is the natural graph-metric
analogue of Muon's polar update. All proofs are in Appendix~\ref{app:theory}.

For the realized GNN module $W\mapsto SHW$, the covariance
$C_g=H^\top S^\top SH/n$ is the exact pullback of the normalized
output-space geometry, whereas $C_H=H^\top H/n$ describes the
pre-propagation map $W\mapsto HW$.
Thus, graph propagation changes the induced geometry through
$S^\top S$, while the underlying module-aware principle may also
extend beyond GNNs; further theoretical motivation is provided
in Appendix~E.

\textbf{Representation-level perturbation.}
The key motivation is that, in GNNs, update scale should be measured by the
perturbation induced after graph propagation, not by parameter-space magnitude
alone. Let $D$ denote the update direction and $\alpha=\eta\, r(d_\ell,d_{\ell+1})$
the scalar multiplying $D$ in Eq.~\eqref{eq:graphmuon_update}. The quantity
governing how an update perturbs a layer's pre-activation is
$\tfrac{1}{\sqrt{n}}\|X_g\Delta W\|_F$, not the parameter-space norm
$\|\Delta W\|_F$. For a vanilla Muon step $\Delta W_{\mathrm{Muon}}=-\alpha D$,
\begin{equation}
    \frac{1}{\sqrt{n}}\left\|X_g \Delta W_{\mathrm{Muon}}\right\|_F
    \;\le\;
    \alpha\sqrt{\lambda_{\max}(C_g)}\,\|D\|_F.
    \label{eq:muon_graph_perturbation}
\end{equation}
Thus, even a matrix-normalized Muon direction can be amplified by the dominant
eigenvalue of the graph-induced covariance.

\textbf{Graph-aware step control.}
The normalization in Eq.~\eqref{eq:graph_scale} removes this amplification.

\begin{theorem}[Graph-representation perturbation control]
\label{thm:graph_perturbation_control}
For any nonzero direction $D$, the \method{} step
$\Delta W_{\mathrm{Graph}}=-\alpha D/s_g$ satisfies:
\begin{equation}
    \frac{1}{\sqrt{n}}\left\|X_g \Delta W_{\mathrm{Graph}}\right\|_F
    \;\le\;
    \alpha \|D\|_F .
    \label{eq:graphmuon_rep_control}
\end{equation}
If $\epsilon_{\mathrm{scale}}=0$ and $X_gD\neq 0$, equality holds.
\end{theorem}

\begin{figure}[t]
    \centering
    \includegraphics[width=0.92\linewidth]{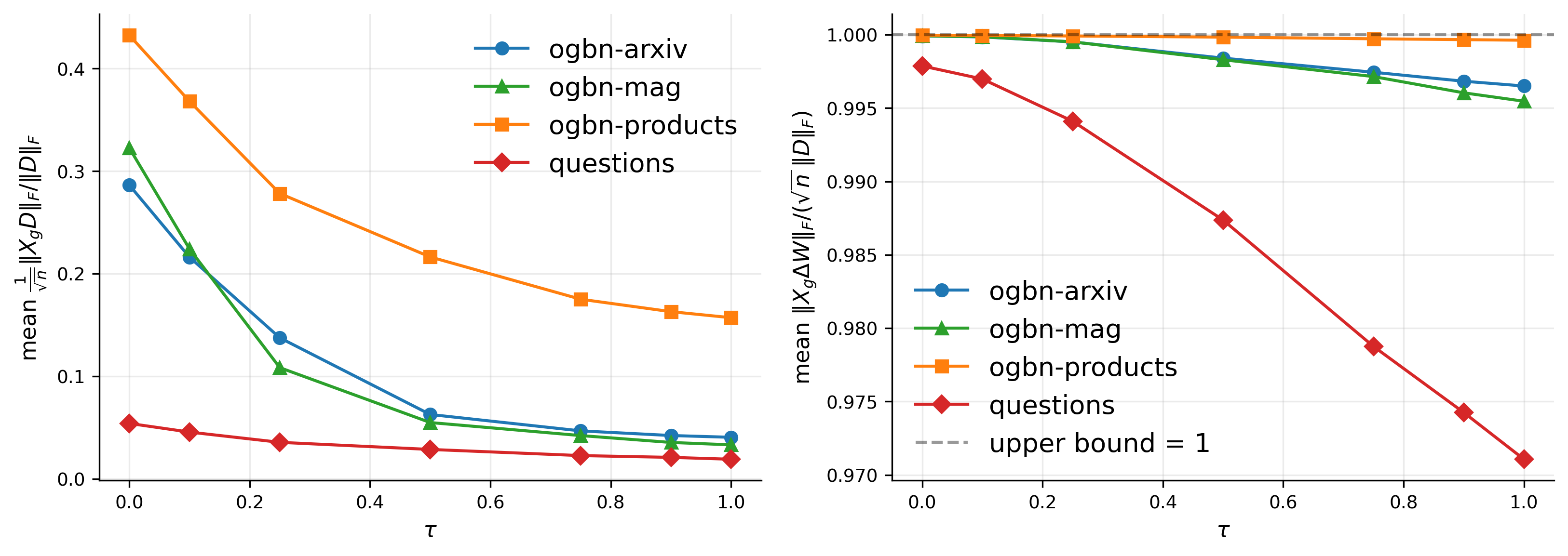}
    \caption{
    Empirical validation of GraphMuon's representation-level perturbation
    control.
    \textbf{Left}: before graph-aware normalization, Euclidean-orthogonalized directions induce
    graph-dependent perturbations whose magnitude varies across datasets and
    values of $\tau$.
    \textbf{Right}: after scaling by $s_g$, the normalized perturbation is tightly
    controlled near the theoretical bound from
    Theorem~\ref{thm:graph_perturbation_control}, confirming that GraphMuon
    calibrates updates by their effect on $X_g$ rather than by Euclidean
    parameter-space scale alone.
    }
    \label{fig:thm1_perturbation_control}
\end{figure}

Unlike Muon, which normalizes in Euclidean matrix space, \method{} rescales
each direction by its induced effect on the graph-propagated representations,
so two directions with equal Frobenius norm produce explicitly calibrated
layer-level perturbations, as empirically validated in
Figure~\ref{fig:thm1_perturbation_control}.

\begin{wrapfigure}{r}{0.48\linewidth}
    \vspace{-0.5em}
    \centering
    \includegraphics[width=0.99\linewidth]{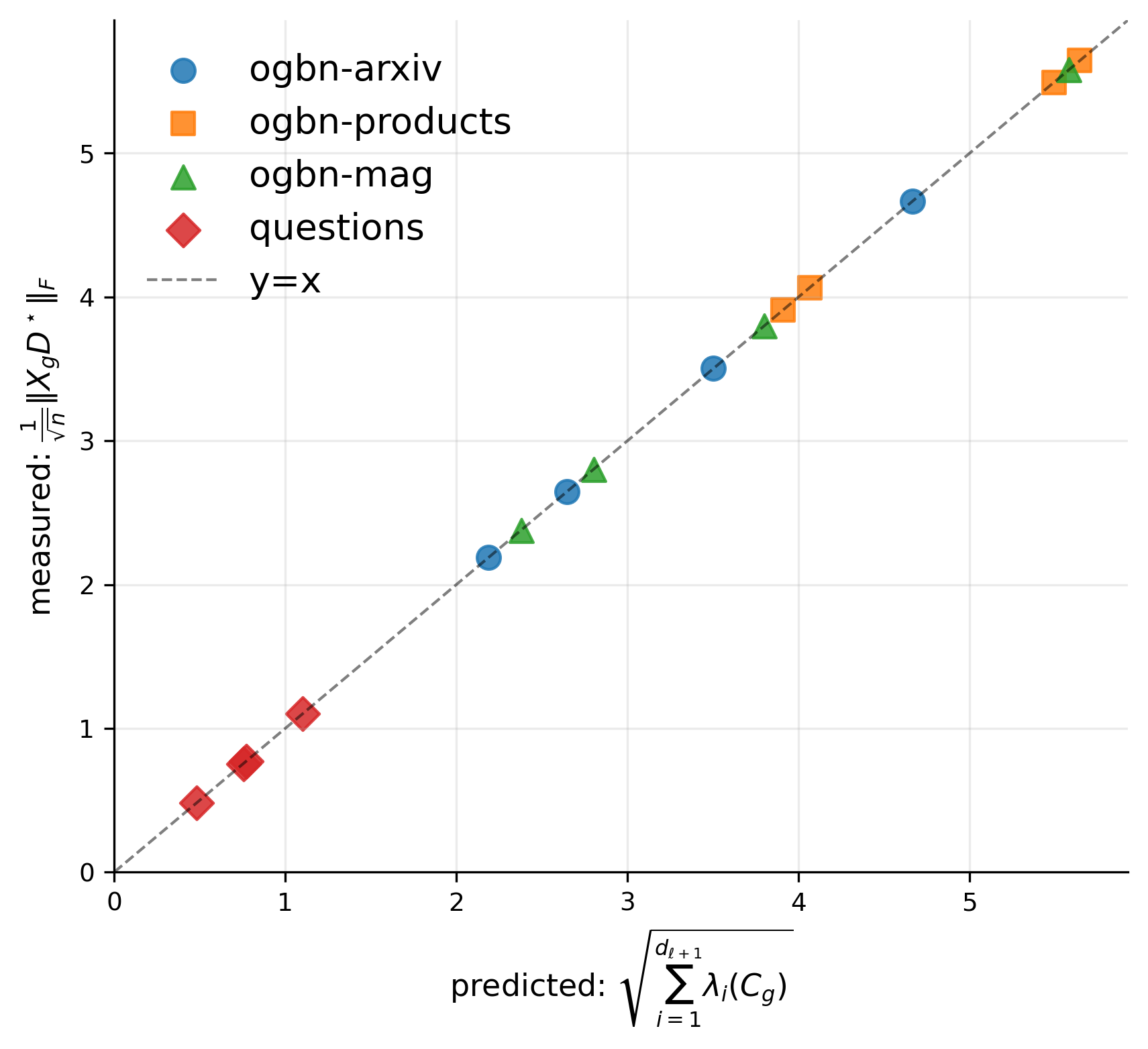}
    \caption{
    Empirical sharpness of Muon's graph-induced amplification.
    For each dataset and layer, we construct the polar direction
    $D^\star$ from the top eigenspaces of the raw graph-induced covariance
    $C_g=X_g^\top X_g/n$.
    The measured amplification,
    $\|X_gD^\star\|_F/\sqrt{n}$, matches the spectral prediction
    $\big(\sum_{i=1}^{d_{\ell+1}}\lambda_i(C_g)\big)^{1/2}$.
    The near-diagonal agreement confirms that the amplification term in
    Proposition~\ref{prop:muon_sharpness} is attainable on real
    graph-propagated representations, rather than a loose
    worst-case bound.
    }
    \label{fig:muon_sharpness}
    \vspace{-6em}
\end{wrapfigure}

\textbf{Sharpness of Muon's graph amplification.}
The $\sqrt{\lambda_{\max}(C_g)}$ factor in Eq.~\eqref{eq:muon_graph_perturbation}
is not a loose artifact of the analysis: there exist valid polar directions for
which it is attained up to dimension-dependent constants.

\begin{proposition}[Sharpness of Muon graph amplification]
\label{prop:muon_sharpness}
Assume $d_\ell\ge d_{\ell+1}$, and let $\lambda_1\ge\cdots\ge\lambda_{d_\ell}$
be the eigenvalues of $C_g$. There exists a matrix $D^\star$ with orthonormal
columns (a feasible polar-factor direction) such that:
\begin{equation}
    \frac{1}{\sqrt{n}}\left\|X_gD^\star\right\|_F
    =
    \sqrt{\sum_{i=1}^{d_{\ell+1}}\lambda_i}
    \;\ge\;
    \sqrt{\lambda_{\max}(C_g)}.
\end{equation}
When $d_{\ell+1}=1$, the $\sqrt{\lambda_{\max}(C_g)}$ amplification is
attained exactly.
\end{proposition}

Proposition~\ref{prop:muon_sharpness} shows that Muon's Euclidean polar
constraint alone does not prevent graph-amplified representation updates.
Such directions are feasible under Muon's geometry and are not explicitly
controlled, motivating the graph-aware normalization in \method{}.

Figure~\ref{fig:muon_sharpness} therefore supports the central motivation for
GraphMuon: controlling the Euclidean geometry of a matrix update does not
necessarily control its induced perturbation after graph propagation.

\textbf{Graph-metric orthogonalization.}
The preceding results focus on update scale. We now justify the
graph-aware orthogonalization step in Eq.~\eqref{eq:graph_aware_orthogonalization}.
Vanilla Muon can be interpreted as applying a polar projection in Euclidean
matrix geometry. \method{} replaces this with the corresponding polar projection
under the graph-induced metric. We analyze the ideal
exact-polar version of the update, replacing the Newton--Schulz approximation
with the exact polar factor \(P(\cdot)\).

\begin{theorem}[Graph-metric polar optimality]
\label{thm:graph_metric_polar}
Assume \(d_\ell\ge d_{\ell+1}\), \(\bar C_g\succ 0\), and
\(\bar C_g^{\tau/2}M\) has full column rank. The exact graph-aware direction
    $D_{g,\tau}
    =
    \bar C_g^{-\tau/2}
    P\!\left(\bar C_g^{\tau/2}M\right)$
satisfies
\begin{equation}
    D_{g,\tau}^\top \bar C_g^\tau D_{g,\tau}
    =
    I,
\end{equation}
and solves the graph-metric Procrustes problem
\begin{equation}
    D_{g,\tau}
    =
    \arg\max_{D^\top\bar C_g^\tau D=I}
    \operatorname{tr}
    \left(
        D^\top \bar C_g^\tau M
    \right).
\end{equation}
\end{theorem}


Theorem~\ref{thm:graph_metric_polar} shows that the graph-aware direction is
not an ad hoc modification of Muon. It is the polar/Muon update obtained after
replacing Euclidean orthogonality by graph-metric orthogonality. \(\tau\)
interpolates between vanilla Muon at \(\tau=0\) and graph-metric
orthogonalization as \(\tau\) increases.

\textbf{Main Takeaways.}
Our theoretical results show that \method{} modifies Muon in two principled
ways. Step normalization controls the representation-level perturbation
\(X_g\Delta W\), while graph-aware orthogonalization computes a polar
direction under the graph-induced metric. In contrast, vanilla Muon balances
updates in Euclidean parameter space and does not explicitly account for the
graph-propagated geometry seen by each GNN layer. The complete layer-wise update is provided in Appendix~\ref{app:full_algorithm}, with additional empirical analyses of the theoretical properties in Appendix~\ref{app:additional_diagnostics}.

\section{Experiments}
\label{sec:experiments}









In this section, we conduct an extensive set of experiments to demonstrate the effectiveness of \textsc{GraphMuon} as an optimizer that adapts GNN updates to graph-propagated representations.
\\
Concretely, our experiments address the following questions:

\textbf{(Q1) Does graph-conditioned optimization improve over parameter-space optimization?}
We compare \textsc{GraphMuon} with AdamW, Muon, Shampoo, and graph-aware natural-gradient baselines to test whether adapting update direction and scale to the graph-propagated covariance improves the main optimization outcome: quicker optimization that maintains competitive or superior test accuracy. This question captures the central distinction of \textsc{GraphMuon}: it does not merely change the optimizer family, but changes the geometry in which GNN weight updates are measured.
\\
\textbf{(Q2) Which component of \textsc{GraphMuon} is responsible for the gains?}
We ablate graph-aware orthogonalization and graph-aware step normalization to determine if  improvements come from changing the update direction, controlling the representation-level step size, or their combination.
\\
\textbf{(Q3) Do empirical diagnostics support the theoretical motivation?}
We measure representation-level perturbations and graph-induced spectral quantities to test whether Euclidean-normalized matrix updates can indeed be amplified by graph-propagated covariance, and if \textsc{GraphMuon} controls it. 


\subsection{Experimental Setup}
\label{subsec:experimental_setup}

\textbf{Datasets.}
We evaluate GraphMuon across nine graph-learning benchmarks,
covering transductive node classification, heterophilous node
classification, inductive graph classification, and graph
regression. The node-classification datasets include
\textsc{ogbn-arxiv}, \textsc{ogbn-products}, and
\textsc{ogbn-mag}~\citep{hu2020open}, together with
\textsc{Questions}~\citep{platonov2023critical},
Cora, and Citeseer. To assess generalization beyond
transductive node prediction, we additionally evaluate
on the molecular graph-classification benchmarks
\textsc{ogbg-molbace} and \textsc{ogbg-moltox21},
and the graph-regression benchmark ZINC.
Additional dataset information is provided in Appendix~G.1.

\textbf{Backbones.}
We evaluate multiple GNN architectures, including GCN,
GAT~\citep{velivckovic2017graph}, GINE, and
GraphGPS~\citep{rampavsek2022recipe}.
GCN provides the primary comparison on the large-scale
OGB node-classification benchmarks. GAT and GraphGPS
extend the evaluation to learned attention-based propagation,
while GINE is used for molecular graph-level tasks.
Additional experiments examine GraphGCN and deeper GNNs.
Within each dataset--architecture configuration, all
optimizers use the same backbone, training budget, and
evaluation protocol, isolating the optimizer as the
varying component. Architectural details are provided
in Appendix~G.2.

\textbf{Baselines.}
The primary validation-loss time-to-target comparison
includes AdamW~\citep{loshchilov2017decoupled} and
Muon~\citep{jordan2024muon}.
AdamW is a strong adaptive first-order baseline with
decoupled weight decay, while Muon is the closest
matrix-aware baseline, applying approximate orthogonalization
to momentum updates. We also consider structured
preconditioning with Shampoo~\citep{gupta2018shampoo}
and graph-aware natural-gradient optimization
~\citep{izadi2020optimization}, with implementation
details provided in Appendix~G.3.

\textbf{Evaluation.}
We follow the corresponding benchmark evaluation protocols
and report predictive performance using test accuracy
for node classification, ROC-AUC for molecular
classification, and MAE for graph regression.
For optimization efficiency, we adopt a fixed
validation-loss time-to-target protocol. For each
dataset--architecture configuration, the target is
the minimum mean validation loss attained by AdamW
across seeds. We measure the first epoch and
corresponding cumulative wall-clock training time
at which each optimizer's mean validation-loss trajectory
reaches this target. Wall-clock speedup is defined as
the ratio of AdamW's time-to-target to that of the
compared optimizer. Configurations that do not reach
the target within the training budget are marked
as ``--''. Additional evaluation results and
per-configuration timings are provided in Appendix~G.

\subsection{Main Results}
\label{subsec:optimizer_comparison_results}

\begin{table*}[t]
\centering
\caption{Validation-loss time-to-target across graph-learning
tasks. The target is the minimum mean validation loss
achieved by AdamW for each dataset--architecture configuration.
Entries report epochs / cumulative wall-clock seconds to
reach the target. Speedup is relative to AdamW; higher is
better. Bold indicates the fastest optimizer. ``--''
denotes failure to reach the target within the training budget.}
\label{tab:main_results}
\small
\setlength{\tabcolsep}{5pt}
\resizebox{\textwidth}{!}{%
\begin{tabular}{llrcccc}
\toprule
\textbf{Dataset} &
\textbf{Backbone} &
\textbf{Target loss} &
\textbf{AdamW (ep / s)} &
\textbf{Muon (ep / s)} &
\textbf{GraphMuon (ep / s)} &
\textbf{Speedup} \\
\midrule
\multicolumn{7}{l}{\emph{Large-scale node classification}}\\
\textsc{ogbn-arxiv} & GCN &
0.8753 & 433 / 30.3 & 188 / 16.6 &
\textbf{164 / 13.1} & \textbf{$2.32\times$} \\

\textsc{ogbn-mag} & GCN &
2.6930 & 100 / 58.3 & 65 / 38.1 &
\textbf{41 / 24.1} & \textbf{$2.42\times$} \\

\textsc{ogbn-products} & GCN &
0.3182 & 300 / 1663.5 & -- &
\textbf{226 / 1257.3} & \textbf{$1.32\times$} \\
\midrule
\multicolumn{7}{l}{\emph{Heterophilous node classification}}\\
\textsc{Questions} & GAT &
0.1217 & 481 / 15.2 & 320 / 10.2 &
\textbf{167 / 7.1} & \textbf{$2.14\times$} \\
\midrule
\multicolumn{7}{l}{\emph{Inductive graph classification}}\\
\textsc{ogbg-molbace} & GINE &
0.6212 & 65 / 59.1 & -- &
\textbf{22 / 29.8} & \textbf{$1.98\times$} \\

\textsc{ogbg-moltox21} & GINE &
0.2946 & 9 / 22.3 & 6 / 22.6 &
\textbf{4 / 18.6} & \textbf{$1.20\times$} \\
\midrule
\multicolumn{7}{l}{\emph{Graph regression}}\\
ZINC & GINE &
0.2502 & 293 / 946.6 & 242 / 1038.4 &
\textbf{106 / 679.0} & \textbf{$1.39\times$} \\
\bottomrule
\end{tabular}%
}
\end{table*}

For the primary comparison, we report validation-loss
time-to-target across the large-scale OGB benchmarks,
heterophilous node classification, and inductive
graph-level tasks. Table~\ref{tab:main_results} shows
that GraphMuon reaches AdamW's validation-loss target
faster across all seven reported configurations.

\begin{table}[t]
\centering
\caption{Generalization across GNN architectures and tasks.
Results are test performance (mean $\pm$ standard deviation
over five seeds). Higher is better for accuracy and ROC-AUC;
lower is better for MAE. Bold indicates the best result.}
\label{tab:architecture_generalization}
\footnotesize
\setlength{\tabcolsep}{4pt}
\begin{tabular}{llccc}
\toprule
\textbf{Dataset / Backbone} &
\textbf{Metric} &
\textbf{AdamW} &
\textbf{Muon} &
\textbf{GraphMuon} \\
\midrule
ogbn-arxiv / GAT &
Acc. $\uparrow$ &
$\mathbf{.6928\pm.0027}$ &
$.6911\pm.0021$ &
$.6926\pm.0027$ \\

ogbn-mag / GraphGPS &
Acc. $\uparrow$ &
$.2939\pm.0018$ &
$.3250\pm.0032$ &
$\mathbf{.3296\pm.0050}$ \\

Tox21 / GINE &
ROC-AUC $\uparrow$ &
$.6633\pm.0223$ &
$.6553\pm.0074$ &
$\mathbf{.6644\pm.0175}$ \\

BACE / GINE &
ROC-AUC $\uparrow$ &
$.7127\pm.0570$ &
$.7810\pm.0194$ &
$\mathbf{.7927\pm.0429}$ \\

ZINC / GINE &
MAE $\downarrow$ &
$.2303\pm.01742$ &
$.1929\pm.00213$ &
$\mathbf{.1897\pm.00112}$ \\
\bottomrule
\end{tabular}
\end{table}

On the three large-scale OGB benchmarks, GraphMuon
achieves wall-clock speedups of $2.32\times$ on
\textsc{ogbn-arxiv}, $2.42\times$ on \textsc{ogbn-mag},
and $1.32\times$ on \textsc{ogbn-products}.
The corresponding 95\% bootstrap confidence intervals
are $[2.01,2.69]$, $[2.29,2.53]$, and $[1.19,1.37]$,
respectively. These results demonstrate faster convergence
to a fixed validation-loss target rather than to a
potentially variable accuracy threshold.

The improvements also extend beyond transductive GCNs.
On \textsc{Questions} with GAT, GraphMuon achieves a
$2.14\times$ wall-clock speedup. On the inductive
molecular graph-classification benchmarks
\textsc{ogbg-molbace} and \textsc{ogbg-moltox21},
it achieves $1.98\times$ and $1.20\times$ speedups,
respectively. On ZINC with GINE, GraphMuon reaches
the validation-loss target $1.39\times$ faster than
AdamW, demonstrating applicability to graph regression.

Additional experiments with GAT, GINE, GraphGPS, and
GraphGCN are reported in
Appendix~\ref{app:architecture_results}.
Across all 15 dataset--architecture configurations,
GraphMuon reaches the target in fewer epochs than AdamW
in every case and achieves the lowest wall-clock
time-to-target in 13 of 15 cases, with an arithmetic-mean
wall-clock speedup of $4.62\times$.
These findings indicate that adapting the optimizer
to graph-propagated representation geometry improves
optimization efficiency across different architectures
and graph-learning tasks.

\begin{figure}[t]
    \centering
    \includegraphics[width=0.425\linewidth]{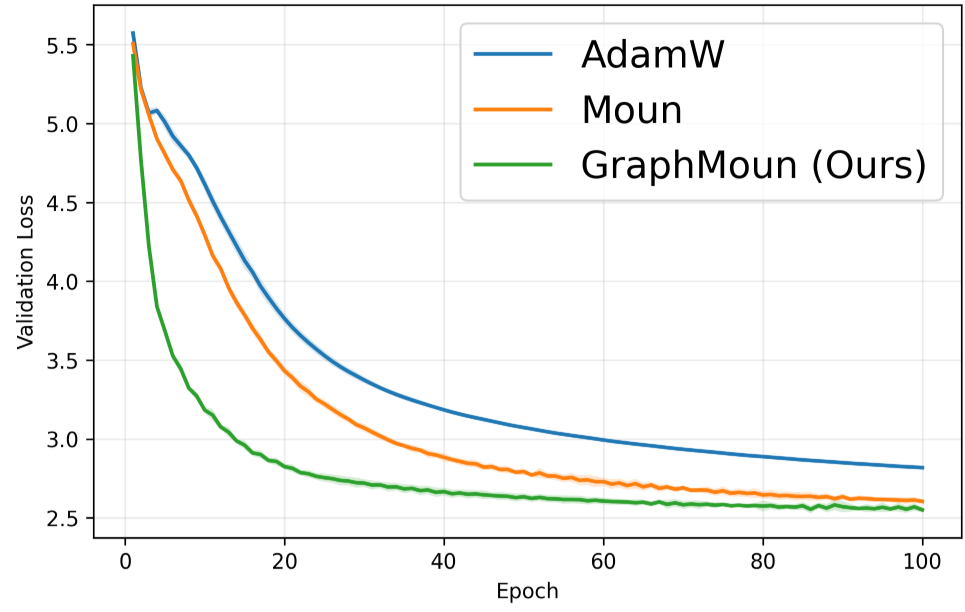}
    \hfill
    \includegraphics[width=0.425\linewidth]{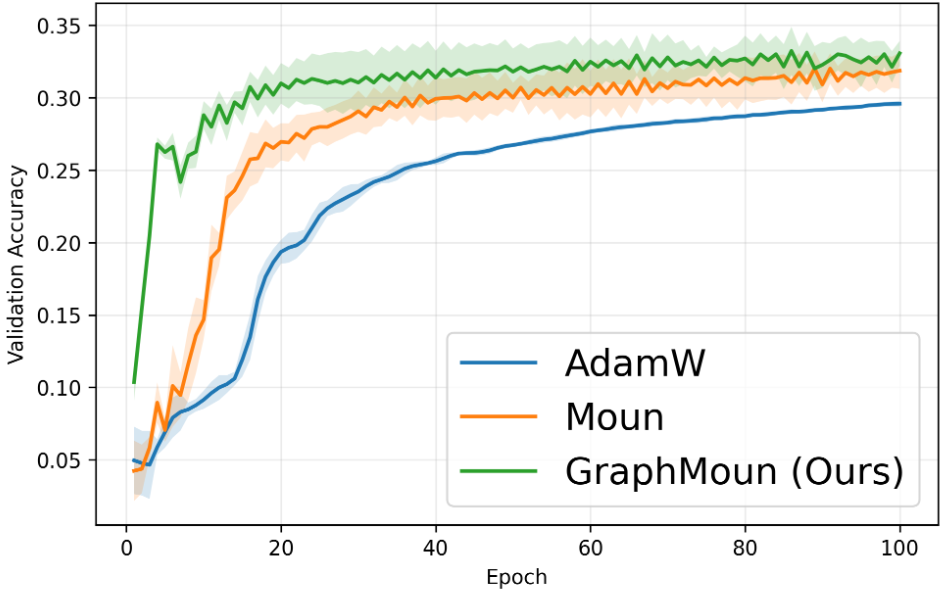}
    \caption{
    Training dynamics on OGBN-MAG. The left panel shows validation loss and the right panel shows validation accuracy over training. \textsc{GraphMuon} makes faster progress in both metrics: it reduces validation loss earlier and reaches strong validation accuracy in fewer epochs than AdamW and Muon, suggesting that graph-induced update geometry can accelerate GNN learning. 
    }
    \label{fig:training_dynamics}
\end{figure}

Overall, \method{} validates our central understanding: normalizing updates in the
graph-propagated layer geometry improves GNN time-to-target without sacrificing
competitive final accuracy. Figure~\ref{fig:training_dynamics} further shows
that these gains come from faster optimization dynamics.

\begin{table}[t]
\small
\setlength{\tabcolsep}{4pt}
\centering
\caption{
Ablation of GraphMuon components on OGB benchmarks. Speedup is measured
relative to vanilla Muon under a fixed rounded-down Muon validation target;
higher is better.
}
\label{tab:ablation}
\begin{tabular}{lcccc}
\toprule
Dataset$\downarrow$ / Method $\rightarrow$ & Muon & + Graph-aware Orthog. & + Graph-aware Step Norm. & GraphMuon (Ours) \\
\midrule
ogbn-arxiv
& $1.00\times$
& $5.39 \pm 0.31\times$
& $8.34 \pm 0.85\times$
& $\mathbf{9.02 \pm 0.80\times}$ \\

ogbn-mag
& $1.00\times$
& $1.07 \pm 0.04\times$
& $2.28 \pm 0.22\times$
& $\mathbf{3.60 \pm 0.12\times}$ \\

ogbn-products
& $1.00\times$
& $1.03 \pm 0.02\times$
& $1.51 \pm 0.05\times$
& $\mathbf{1.90 \pm 0.06\times}$ \\
\bottomrule
\end{tabular}
\end{table}

\textbf{Generalization across architectures and tasks.}
To evaluate generalization beyond GCN-based node classification,
we extend the evaluation to GAT, GINE, and GraphGPS, covering
node classification, inductive graph classification, and graph
regression. As shown in Table~\ref{tab:architecture_generalization},
GraphMuon achieves the best numerical performance in four of
five configurations and remains competitive with AdamW on
GAT. The improvements on GINE and GraphGPS demonstrate
that the method also applies to graph-level prediction and
learned attention-based architectures. Additional validation-loss
time-to-target results across 15 configurations are provided
in Appendix~\ref{app:architecture_results}.

\subsection{Ablation Study}
\label{subsec:ablation}

We ablate the two main components of GraphMuon: graph-aware orthogonalization
and graph-aware step normalization. To focus on optimization speed, we use a
relaxed time-to-target metric. For each dataset, the target is the best
validation loss reached by vanilla Muon, rounded down to two decimal places
(e.g., \(0.723 \mapsto 0.72\)). This threshold is close to the converged Muon
performance while avoiding sensitivity to small validation fluctuations. We then
measure the wall-clock time required to first reach this threshold and report
speedup relative to vanilla Muon.
\\
Table~\ref{tab:ablation} reports the main ablation on the OGB benchmarks, where the relaxed target is
not saturated and therefore distinguishes the contribution of each component.
The \textsc{questions} ablation is omitted from the main table because all
graph-aware variants reach the rounded Muon target after a single optimization
step, making the time-to-target metric uninformative for component comparison.

\section{Conclusions}
\label{sec:discussion_limitations}

We motivate our GraphMuon by a representation-space view of optimization in GNNs.
For a layer with graph-propagated input $X_g=SH$, the effective update size is
not $\|\Delta W\|_F$, but the induced perturbation
$\|X_g\Delta W\|_F/\sqrt{n}$. Our analysis shows that a Euclidean-normalized
Muon update can still be amplified by the graph-induced covariance
$C_g=X_g^\top X_g/n$, through its dominant eigenspaces. GraphMuon addresses this
by normalizing the update according to its induced representation-level effect
and by replacing Euclidean matrix geometry with a graph-metric analogue.

This theory also explains when GraphMuon may provide less benefit. The method is
most useful when $C_g$ is stable, anisotropic, and informative. On very small
graphs, however, $C_g$ may be low-rank and noisy, especially when the number of
nodes is small relative to the hidden dimension. In this regime, the
graph-induced metric can be poorly estimated, and controlling
$\|X_g\Delta W\|_F$ does not necessarily imply faster convergence or better
generalization. AdamW can therefore remain competitive, since its coordinate-wise
moment estimates do not rely on estimating a graph-level covariance.

This limitation is particularly relevant for small inductive graph-level tasks,
where each graph contains few node representations. In such settings, the main
optimization challenge may be dominated by sparse local features, atom or bond
types, and mini-batch variability rather than by graph-induced update
amplification. A natural direction for future work is to adapt the strength of
graph awareness to the reliability of $C_g$, for example, using effective rank,
spectral anisotropy, or covariance stability to interpolate between vanilla Muon and fully graph-aware updates.

\section*{Acknowledgments and Disclosures}
 ME acknowledges support from the the Israel Science Foundation through a personal grant (ISF 3062/26) and an
equipment grant (ISF 3066/26), as well as support from the Israeli Ministry of Innovation, Science \& Technology.
The authors declare no competing interests.


\bibliographystyle{plainnat}
\bibliography{iclr2025_conference}

@article{robbins1951stochastic,
  title={A stochastic approximation method},
  author={Robbins, Herbert and Monro, Sutton},
  journal={The annals of mathematical statistics},
  pages={400--407},
  year={1951},
  publisher={JSTOR}
}

@inproceedings{eliasof2025graph,
  title={Graph adaptive autoregressive moving average models},
  author={Eliasof, Moshe and Gravina, Alessio and Ceni, Andrea and Gallicchio, Claudio and Bacciu, Davide and Sch{\"o}nlieb, Carola-Bibiane},
  booktitle={Forty-second International Conference on Machine Learning},
  year={2025}
}

@book{horn2012matrix,
  title     = {Matrix Analysis},
  author    = {Horn, Roger A. and Johnson, Charles R.},
  edition   = {2},
  publisher = {Cambridge University Press},
  year      = {2012}
}

@article{kingma2014adam,
  title={Adam: A method for stochastic optimization},
  author={Kingma, Diederik P and Ba, Jimmy},
  journal={arXiv preprint arXiv:1412.6980},
  year={2014}
}

@article{loshchilov2017decoupled,
  title={Decoupled weight decay regularization},
  author={Loshchilov, Ilya and Hutter, Frank},
  journal={arXiv preprint arXiv:1711.05101},
  year={2017}
}

@article{kipf2016semi,
  title={Semi-supervised classification with graph convolutional networks},
  author={Kipf, Thomas N and Welling, Max},
  journal={arXiv preprint arXiv:1609.02907},
  year={2016}
}

@article{bernstein2025manifolds,
  author = {Jeremy Bernstein},
  title = {Modular Manifolds},
  journal = {Thinking Machines Lab: Connectionism},
  year = {2025},
  note = {https://thinkingmachines.ai/blog/modular-manifolds/},
  doi = {10.64434/tml.20250926}
}

@inproceedings{hu2020strategies,
  title={Strategies For Pre-training Graph Neural Networks},
  author={Hu, W and Liu, B and Gomes, J and Zitnik, M and Liang, P and Pande, V and Leskovec, J},
  booktitle={International Conference on Learning Representations (ICLR)},
  year={2020}
}

@article{platonov2023critical,
  title={A critical look at the evaluation of GNNs under heterophily: Are we really making progress?},
  author={Platonov, Oleg and Kuznedelev, Denis and Diskin, Michael and Babenko, Artem and Prokhorenkova, Liudmila},
  journal={arXiv preprint arXiv:2302.11640},
  year={2023}
}

@article{hu2020open,
  title={Open graph benchmark: Datasets for machine learning on graphs},
  author={Hu, Weihua and Fey, Matthias and Zitnik, Marinka and Dong, Yuxiao and Ren, Hongyu and Liu, Bowen and Catasta, Michele and Leskovec, Jure},
  journal={Advances in neural information processing systems},
  volume={33},
  pages={22118--22133},
  year={2020}
}

@article{mantri2024digraf,
  title={Digraf: Diffeomorphic graph-adaptive activation function},
  author={Mantri, Krishna S and Wang, Xinzhi and Sch{\"o}nlieb, Carola-Bibiane and Ribeiro, Bruno and Bevilacqua, Beatrice and Eliasof, Moshe},
  journal={Advances in Neural Information Processing Systems},
  volume={37},
  pages={3649--3681},
  year={2024}
}

@article{rampavsek2022recipe,
  title={Recipe for a general, powerful, scalable graph transformer},
  author={Ramp{\'a}{\v{s}}ek, Ladislav and Galkin, Michael and Dwivedi, Vijay Prakash and Luu, Anh Tuan and Wolf, Guy and Beaini, Dominique},
  journal={Advances in Neural Information Processing Systems},
  volume={35},
  pages={14501--14515},
  year={2022}
}

@article{eliasof2024granola,
  title={Granola: Adaptive normalization for graph neural networks},
  author={Eliasof, Moshe and Bevilacqua, Beatrice and Sch{\"o}nlieb, Carola-Bibiane and Maron, Haggai},
  journal={Advances in Neural Information Processing Systems},
  volume={37},
  pages={90514--90551},
  year={2024}
}

@inproceedings{cai2021graphnorm,
  title={Graphnorm: A principled approach to accelerating graph neural network training},
  author={Cai, Tianle and Luo, Shengjie and Xu, Keyulu and He, Di and Liu, Tie-yan and Wang, Liwei},
  booktitle={International Conference on Machine Learning},
  pages={1204--1215},
  year={2021},
  organization={PMLR}
}

@article{xu2018powerful,
  title={How powerful are graph neural networks?},
  author={Xu, Keyulu and Hu, Weihua and Leskovec, Jure and Jegelka, Stefanie},
  journal={arXiv preprint arXiv:1810.00826},
  year={2018}
}

@inproceedings{velivckovic2018graph,
  title={Graph Attention Networks},
  author={Veli{\v{c}}kovi{\'c}, Petar and Cucurull, Guillem and Casanova, Arantxa and Romero, Adriana and Li{\`o}, Pietro and Bengio, Yoshua},
  booktitle={International Conference on Learning Representations},
  year={2018}
}

@inproceedings{yang2024graph,
  title={How Graph Neural Networks Learn: Lessons from Training Dynamics},
  author={Yang, Chenxiao and Wu, Qitian and Wipf, David and Sun, Ruoyu and Yan, Junchi},
  booktitle={International Conference on Machine Learning},
  pages={56594--56623},
  year={2024},
  organization={PMLR}
}

@inproceedings{xu2021optimization,
  title={Optimization of graph neural networks: Implicit acceleration by skip connections and more depth},
  author={Xu, Keyulu and Zhang, Mozhi and Jegelka, Stefanie and Kawaguchi, Kenji},
  booktitle={International conference on machine learning},
  pages={11592--11602},
  year={2021},
  organization={PMLR}
}

@inproceedings{izadi2020optimization,
  title={Optimization of graph neural networks with natural gradient descent},
  author={Izadi, Mohammad Rasool and Fang, Yihao and Stevenson, Robert and Lin, Lizhen},
  booktitle={2020 IEEE international conference on big data (big data)},
  pages={171--179},
  year={2020},
  organization={IEEE}
}

@inproceedings{martens2015optimizing,
  title={Optimizing neural networks with kronecker-factored approximate curvature},
  author={Martens, James and Grosse, Roger},
  booktitle={International conference on machine learning},
  pages={2408--2417},
  year={2015},
  organization={PMLR}
}

@article{duchi2011adaptive,
  title={Adaptive subgradient methods for online learning and stochastic optimization.},
  author={Duchi, John and Hazan, Elad and Singer, Yoram},
  journal={Journal of machine learning research},
  volume={12},
  number={7},
  year={2011}
}

@inproceedings{gupta2018shampoo,
  title={Shampoo: Preconditioned stochastic tensor optimization},
  author={Gupta, Vineet and Koren, Tomer and Singer, Yoram},
  booktitle={International Conference on Machine Learning},
  pages={1842--1850},
  year={2018},
  organization={PMLR}
}

@misc{jordan2024muon,
  author       = {Keller Jordan and Yuchen Jin and Vlado Boza and Jiacheng You and
                  Franz Cesista and Laker Newhouse and Jeremy Bernstein},
  title        = {Muon: An optimizer for hidden layers in neural networks},
  year         = {2024},
  url          = {https://kellerjordan.github.io/posts/muon/}
}

@inproceedings{fan1955some,
  title={Some metric inequalities in the space of matrices},
  author={Fan, Ky and Hoffman, Alan J and others},
  booktitle={Proc. Amer. Math. Soc},
  volume={6},
  number={1},
  pages={111--116},
  year={1955},
  organization={World Scientific}
}

@article{velivckovic2017graph,
  title={Graph attention networks},
  author={Veli{\v{c}}kovi{\'c}, Petar and Cucurull, Guillem and Casanova, Arantxa and Romero, Adriana and Lio, Pietro and Bengio, Yoshua},
  journal={arXiv preprint arXiv:1710.10903},
  year={2017}
}

@article{paszke2019pytorch,
  title={Pytorch: An imperative style, high-performance deep learning library},
  author={Paszke, Adam and Gross, Sam and Massa, Francisco and Lerer, Adam and Bradbury, James and Chanan, Gregory and Killeen, Trevor and Lin, Zeming and Gimelshein, Natalia and Antiga, Luca and others},
  journal={Advances in neural information processing systems},
  volume={32},
  year={2019}
}

@article{fey2019fast,
  title={Fast graph representation learning with PyTorch Geometric},
  author={Fey, Matthias and Lenssen, Jan Eric},
  journal={arXiv preprint arXiv:1903.02428},
  year={2019}
}

@book{bhatia1997matrix,
  title     = {Matrix Analysis},
  author    = {Bhatia, Rajendra},
  series    = {Graduate Texts in Mathematics},
  volume    = {169},
  year      = {1997},
  publisher = {Springer},
  address   = {New York}
}

@article{scarselli2008graph,
  title= {The {G}raph {N}eural {N}etwork Model},
  author    = {Scarselli, Franco and Gori, Marco and Tsoi, Ah Chung and
               Hagenbuchner, Markus and Monfardini, Gabriele},
  journal   = {IEEE Transactions on Neural Networks},
  volume    = {20},
  number    = {1},
  pages     = {61--80},
  year      = {2009},
  publisher = {IEEE}
}

@inproceedings{gilmer2017neural,
  title     = {Neural Message Passing for Quantum Chemistry},
  author    = {Gilmer, Justin and Schoenholz, Samuel S. and Riley, Patrick F.
               and Vinyals, Oriol and Dahl, George E.},
  booktitle = {Proceedings of the 34th International Conference on Machine Learning},
  pages     = {1263--1272},
  year      = {2017},
  volume    = {70},
  series    = {Proceedings of Machine Learning Research},
  publisher = {PMLR}
}

@inproceedings{hamilton2017inductive,
  title     = {Inductive Representation Learning on Large Graphs},
  author    = {Hamilton, William L. and Ying, Zhitao and Leskovec, Jure},
  booktitle = {Advances in Neural Information Processing Systems},
  volume    = {30},
  pages     = {1024--1034},
  year      = {2017}
}

@article{liu2025muon,
  title={Muon is scalable for llm training},
  author={Liu, Jingyuan and Su, Jianlin and Yao, Xingcheng and Jiang, Zhejun and Lai, Guokun and Du, Yulun and Qin, Yidao and Xu, Weixin and Lu, Enzhe and Yan, Junjie and others},
  journal={arXiv preprint arXiv:2502.16982},
  year={2025}
}

@article{team2025kimi,
  title={Kimi k2: Open agentic intelligence},
  author={Team, Kimi and Bai, Yifan and Bao, Yiping and Charles, Y and Chen, Cheng and Chen, Guanduo and Chen, Haiting and Chen, Huarong and Chen, Jiahao and Chen, Ningxin and others},
  journal={arXiv preprint arXiv:2507.20534},
  year={2025}
}

@article{team2026kimi,
  title={Kimi K2. 5: Visual Agentic Intelligence},
  author={Team, Kimi and Bai, Tongtong and Bai, Yifan and Bao, Yiping and Cai, SH and Cao, Yuan and Charles, Y and Che, HS and Chen, Cheng and Chen, Guanduo and others},
  journal={arXiv preprint arXiv:2602.02276},
  year={2026}
}

@inproceedings{schlichtkrull2018modeling,
  title     = {Modeling Relational Data with Graph Convolutional Networks},
  author    = {Schlichtkrull, Michael and Kipf, Thomas N. and Bloem, Peter and
               van den Berg, Rianne and Titov, Ivan and Welling, Max},
  booktitle = {The Semantic Web:15th International Conference, {ESWC} 2018},
  pages     = {593--607},
  year      = {2018},
  publisher = {Springer}
}

@inproceedings{zhu2020beyond,
  title     = {Beyond Homophily in Graph Neural Networks:
               Current Limitations and Effective Designs},
  author    = {Zhu, Jiong and Yan, Yujun and Zhao, Lingxiao and
               Heimann, Mark and Akoglu, Leman and Koutra, Danai},
  booktitle = {Advances in Neural Information Processing Systems},
  volume    = {33},
  pages     = {7793--7804},
  year      = {2020}
}


\clearpage

\appendix
\section{Background}
\label{app:full_background}
Here we provide complete definitions and implementation details for the
methods referenced in the main paper. We give the standard GCN layer formulation,
the Adam and AdamW update rules, and the full Muon algorithm including the
Newton--Schulz iteration and its practical implementation details.
\subsection{GCN Layer Definition}
\label{app:gcn_details}
Given the symmetrically normalized adjacency $S$ (defined in
Section~\ref{sec:background}), a standard GCN layer computes
\begin{equation}
    H^{(\ell+1)}
    =
    \sigma\!\left(SH^{(\ell)}W^{(\ell)}\right),
\end{equation}
where $W^{(\ell)}\in\mathbb{R}^{d_\ell\times d_{\ell+1}}$ is the trainable
weight matrix and $\sigma(\cdot)$ is a nonlinear activation. The degree matrix
entries are
\[
    \widetilde{D}_{ii}
    =
    \sum_{j=1}^{n} \widetilde{A}_{ij}
    =
    1 + \sum_{j=1}^{n} A_{ij}.
\]
\subsection{Adam and AdamW}
\label{app:adam}
Let $\theta_t$ be the parameters and $g_t = \nabla_\theta \mathcal{L}(\theta_t)$
the stochastic gradient. Adam maintains first and second moment estimates:
\begin{align}
    m_t &= \beta_1 m_{t-1} + (1-\beta_1)g_t, \\
    v_t &= \beta_2 v_{t-1} + (1-\beta_2)(g_t \odot g_t).
\end{align}
With bias correction, the update is
\begin{equation}
    \theta_{t+1}
    =
    \theta_t
    -
    \eta
    \frac{\widehat{m}_t}{\sqrt{\widehat{v}_t}+\epsilon}.
\end{equation}
AdamW decouples weight decay:
\begin{equation}
    \theta_{t+1}
    =
    (1-\eta\lambda)\theta_t
    -
    \eta
    \frac{\widehat{m}_t}{\sqrt{\widehat{v}_t}+\epsilon}.
\end{equation}
Adam-based optimizers operate coordinate-wise, normalizing each parameter
independently, and do not exploit matrix structure or graph-induced geometry.
\subsection{Muon: Full Details}
\label{app:muon_details}
The polar factor of a matrix $M$ is
\begin{equation}
    \polar(M)
    =
    M(M^\top M)^{-1/2},
\end{equation}
with the analogous expression $(MM^\top)^{-1/2}M$ when $d_{\mathrm{in}}<d_{\mathrm{out}}$.
Muon approximates this via Newton--Schulz iterations. Starting from
\begin{equation}
    X_0 = \frac{M_t}{\|M_t\|_F+\epsilon},
\end{equation}
it applies the quintic iteration for $k=0,\dots,K-1$:
\begin{align}
    A_k      &= X_k X_k^\top, \\
    X_{k+1}  &= a X_k + b A_k X_k + c A_k^2 X_k,
\end{align}
with coefficients
\begin{equation}
    a = 3.4445, \qquad b = -4.7750, \qquad c = 2.0315,
\end{equation}
selected via an ad-hoc gradient-based search in the original
implementation~\citep{jordan2024muon} to rapidly approximate the polar factor.
When $d_{\mathrm{in}} > d_{\mathrm{out}}$, $M_t$ is transposed before the
iteration (so the squared matrix is formed over the smaller dimension) and
transposed back afterward, reducing the dominant matrix-multiplication cost.
After $K=5$ iterations the direction is $D_t = \NS_K(M_t)$.
The vanilla Muon update with decoupled weight decay is
\begin{equation}
    W_{t+1} = (1-\eta\lambda)W_t - \eta D_t.
\end{equation}

\section{Additional Related Work}
\label{app:additional_related_work}

\textbf{Graph-adaptive neural components.}
A broad line of work adapts standard neural-network components to
graph-structured data. GCNs use normalized adjacency propagation
\citep{kipf2016semi}, GAT adapts attention to graph neighborhoods
\citep{velivckovic2018graph}, and GIN studies aggregation from the perspective
of expressive power~\citep{xu2018powerful}. Graph-specific design has also been
extended to normalization and modular architectures. GraphNorm introduces
graph-specific normalization~\citep{cai2021graphnorm}, GRANOLA uses
neighborhood-structure-based normalization~\citep{eliasof2024granola},
GraphGPS combines local message passing with global attention and positional
encodings~\citep{rampavsek2022recipe}, GRAMA~\citep{eliasof2025graph} learns a graph-adaptive autoregressive moving average filtering technique, and DiGRAF learns graph-adaptive
activation functions~\citep{mantri2024digraf}. These works support the broader
principle that generic deep-learning components often benefit from
graph-specific adaptation. \method{} applies this principle to optimization
rather than architecture.

\textbf{GNN training dynamics.}
Several works study why GNN optimization differs from optimization in standard
Euclidean architectures. Natural-gradient methods exploit information-geometric
structure in graph-based semi-supervised learning~\citep{izadi2020optimization};
skip connections, depth, and label distributions have been shown to affect GNN
convergence~\citep{xu2021optimization}; and recent training-dynamics analyses
show that gradient descent can implicitly exploit graph structure through
kernel--graph alignment~\citep{yang2024graph}. These results motivate our
representation-space view of optimization: the effect of a weight update in a
GNN is determined not only by the update matrix itself, but by the
graph-propagated representations on which it acts.

\section{\GraphMuon{} Full Algorithm}
\label{app:full_algorithm}

Algorithm~\ref{alg:graphmuon} gives the complete \GraphMuon{} update for
matrix-valued GNN weight matrices. The algorithm follows the update defined in
Section~\ref{sec:method}: for each layer, it computes the graph-propagated input,
forms the corresponding graph-induced covariance, applies graph-aware
orthogonalization to the Muon momentum direction, and rescales the resulting
update by its induced representation-level perturbation. Non-matrix parameters,
such as biases or normalization parameters, can be updated with the accompanying
base optimizer.

\begin{algorithm}[t]
\caption{Full \GraphMuon{} update for GNN weight matrices}
\label{alg:graphmuon}
\begin{algorithmic}[1]
\REQUIRE Learning rate $\eta$, momentum $\mu$, weight decay $\lambda$,
Newton--Schulz steps $K$, graph-awareness strength $\tau$,
regularization constants $\epsilon_{\mathrm{cov}}$ and
$\epsilon_{\mathrm{scale}}$; normalized adjacency $S$;
layer representations $\{H^{(\ell)}\}_\ell$; parameters
$\{W_t^{(\ell)}\}_\ell$; momentum buffers $\{B_{t-1}^{(\ell)}\}_\ell$
\ENSURE Updated parameters $\{W_{t+1}^{(\ell)}\}_\ell$

\FOR{each matrix-valued GNN layer $\ell$}
  \STATE $G_t^{(\ell)} \gets \nabla_{W^{(\ell)}} \mathcal{L}_t$
         \COMMENT{gradient}
  \STATE $B_t^{(\ell)} \gets \mu B_{t-1}^{(\ell)} + G_t^{(\ell)}$
         \COMMENT{momentum buffer}
  \STATE $M_t^{(\ell)} \gets G_t^{(\ell)} + \mu B_t^{(\ell)}$
         \COMMENT{Nesterov momentum matrix}

  \STATE $X_g^{(\ell)} \gets S H^{(\ell)}$
         \COMMENT{graph-propagated layer input}
  \STATE $C_g^{(\ell)} \gets \frac{1}{n} X_g^{(\ell)\top} X_g^{(\ell)}$
         \COMMENT{graph-induced covariance}

  \STATE $\bar{C}_g^{(\ell)} \gets
  C_g^{(\ell)} \big/ \left(\operatorname{tr}(C_g^{(\ell)})/d_\ell\right)
  + \epsilon_{\mathrm{cov}} I$
         \COMMENT{trace-normalize and regularize}

  \STATE $A_{\pm}^{(\ell)} \gets
  \MatPow\!\left(\bar{C}_g^{(\ell)}, \tau/2\right)$
         \COMMENT{covariance half-powers}

  \STATE $D^{(\ell)} \gets
  A_{-}^{(\ell)} \NS_K\!\left(A_{+}^{(\ell)} M_t^{(\ell)}\right)$
         \COMMENT{graph-aware Muon direction}

  \STATE $s_g^{(\ell)} \gets
  \sqrt{
  \operatorname{tr}\!\left(D^{(\ell)\top} C_g^{(\ell)} D^{(\ell)}\right)
  \big/ \|D^{(\ell)}\|_F^2
  + \epsilon_{\mathrm{scale}}
  }$
         \COMMENT{graph-aware scale}

  \STATE $\alpha_\ell \gets \eta\, r(d_\ell,d_{\ell+1})$
         \COMMENT{shape-corrected layerwise step size}

  \STATE $W_{t+1}^{(\ell)} \gets
  \left(1-\alpha_\ell\lambda\right) W_t^{(\ell)}
  - \alpha_\ell D^{(\ell)} / s_g^{(\ell)}$
         \COMMENT{decoupled weight decay and update}
\ENDFOR

\RETURN $\{W_{t+1}^{(\ell)}\}_\ell$
\end{algorithmic}
\end{algorithm}

\section{Step Normalization: Use of Raw vs.\ Normalized Covariance}
\label{app:step_norm_derivation}
The trace-normalized covariance $\bar{C}_g^{(\ell)}$ used in
Eq.~\eqref{eq:graph_aware_orthogonalization} removes the layer-wise average
feature energy so that orthogonalization depends only on the \emph{anisotropy}
of the graph-propagated features. Step normalization has the opposite goal: it
should measure the actual representation-level perturbation induced by the
update. Since
\[
    \operatorname{tr}\!\left(D^{(\ell)\top} C_g^{(\ell)} D^{(\ell)}\right)
    =
    \frac{1}{n}
    \left\|X_g^{(\ell)}D^{(\ell)}\right\|_F^2,
\]
using the raw $C_g^{(\ell)}$ makes $s_g^{(\ell)}$ equal (up to $\epsilon_{\mathrm{scale}}$)
to the RMS pre-activation perturbation per column of $X_g^{(\ell)}D^{(\ell)}$.
Substituting $\bar{C}_g^{(\ell)}$ would divide out the average feature energy
and remove precisely the layer-wise scale that the normalization is intended to
control.

\section{Proofs of Theoretical Results}
\label{app:theory}

Throughout this appendix we drop the layer superscript $(\ell)$ and write
$\alpha = \eta\,r(d_\ell, d_{\ell+1})$ for the effective per-layer step size.
We denote by $d_{\mathrm{in}} = d_\ell$ and $d_{\mathrm{out}} = d_{\ell+1}$
the input and output dimensions of the weight matrix $W \in \R^{d_{\mathrm{in}}
\times d_{\mathrm{out}}}$.

\subsection{Shared Setup}
\label{app:setup}

\textbf{Graph-induced covariance.}
Given graph-propagated features $X_g = SH \in \R^{n \times d_{\mathrm{in}}}$,
define the graph-induced covariance
\begin{equation}
  C_g \;=\; \frac{1}{n}\,X_g^\top X_g \;\in\; \R^{d_{\mathrm{in}} \times d_{\mathrm{in}}}.
  \label{eq:app_cov_def}
\end{equation}
$C_g$ is positive semidefinite.  Let $\lambda_1 \ge \lambda_2 \ge \cdots \ge
\lambda_{d_{\mathrm{in}}} \ge 0$ denote its eigenvalues and $V = [v_1, \ldots,
v_{d_{\mathrm{in}}}]$ the corresponding orthonormal eigenvectors, so that
$C_g = V\Lambda V^\top$ with $\Lambda = \mathrm{diag}(\lambda_1,\ldots,
\lambda_{d_{\mathrm{in}}})$.

\textbf{Key identity.}
For any matrix $\Delta W \in \R^{d_{\mathrm{in}} \times d_{\mathrm{out}}}$,
\begin{align}
  \frac{1}{n}\bigl\|X_g \Delta W\bigr\|_F^2
  &= \frac{1}{n}\operatorname{tr}\!\bigl(\Delta W^\top X_g^\top X_g\,\Delta W\bigr)
   = \operatorname{tr}\!\bigl(\Delta W^\top C_g\,\Delta W\bigr).
  \label{eq:app_trace_id}
\end{align}

\textbf{Spectral bound.}
For any positive semidefinite $C$ and any matrix $V$ with columns $v_j$,
\begin{equation}
  \operatorname{tr}(V^\top C V)
  = \sum_j v_j^\top C v_j
  \;\le\; \lambda_{\max}(C)\sum_j \|v_j\|_2^2
  = \lambda_{\max}(C)\,\|V\|_F^2.
  \label{eq:app_spectral}
\end{equation}

\textbf{Module-induced geometry.}
Consider the realized GNN module $F_g(W)=SHW$, with $S$ and
$H$ fixed at the current step. Its Jacobian maps a parameter
perturbation $\Delta W$ to $SH\Delta W$. Consequently, the
normalized Frobenius geometry of the module output induces
the parameter-space inner product
\begin{equation}
\frac{1}{n}\langle SH\Delta W_1,SH\Delta W_2\rangle_F
=\operatorname{tr}(\Delta W_1^\top C_g\Delta W_2),
\qquad
C_g=\frac{1}{n}H^\top S^\top SH.
\end{equation}
In contrast, the pre-propagation map $W\mapsto HW$ induces
$C_H=H^\top H/n$, with
\begin{equation}
C_g-C_H=\frac{1}{n}H^\top(S^\top S-I)H.
\end{equation}
The two metrics coincide when propagation is isometric on
the feature subspace, including $S=I$. Otherwise, $C_H$
does not capture the amplification or attenuation introduced
by graph propagation.

\textbf{Connection to the optimization objective.}
Let $Z=SHW$, and assume the loss $\mathcal{L}$ is locally
$\beta$-smooth in $Z$ under the normalized output norm
$\|U\|_{\mathrm{out}}=\|U\|_F/\sqrt{n}$. Then
\begin{equation}
\begin{aligned}
\mathcal{L}(Z+SH\Delta W)
\leq {}& \mathcal{L}(Z)
+\langle\nabla_Z\mathcal{L},SH\Delta W\rangle_F\\
&+\frac{\beta}{2}
\operatorname{tr}(\Delta W^\top C_g\Delta W).
\end{aligned}
\end{equation}
Thus, $\|SH\Delta W\|_F/\sqrt{n}$ is precisely the
perturbation magnitude governing this local smoothness
bound. A parameter-space norm alone cannot distinguish
directions with different effects on the realized module:
for any unit eigenvector $v_i$ of $C_g$ with eigenvalue
$\lambda_i$ and any unit vector $q$, the perturbation
$\Delta W_i=v_iq^\top$ satisfies
$\|\Delta W_i\|_2=1$, while
$\|SH\Delta W_i\|_F^2/n=\lambda_i$.
This motivates controlling representation-level
perturbations rather than relying exclusively on
Euclidean normalization of the parameter update.

\textbf{Classical orthogonal Procrustes result.}
\begin{lemma}[Orthogonal Procrustes, \citep{fan1955some}]
\label{lem:procrustes}
Let $A \in \R^{m \times k}$ with $m \ge k$ and $A = U\Sigma V^\top$ its
(thin) SVD.  Then
\begin{equation}
  \max_{\substack{Q \in \R^{m \times k} \\ Q^\top Q = I}}
  \operatorname{tr}(Q^\top A)
  = \operatorname{tr}(\Sigma),
\end{equation}
and the unique maximiser (assuming $A$ has full column rank) is the polar
factor $P(A) = UV^\top$.
\end{lemma}

\noindent This classical result (see also \citep{bhatia1997matrix}) is used
in the proof of Theorem~\ref{thm:graph_metric_polar}.

\subsection{Proof of Theorem~\ref{thm:graph_perturbation_control}
           (Graph-Representation Perturbation Control)}
\label{app:proof_thm1}

\begin{proof}
The \method{} step is $\Delta W_{\mathrm{Graph}} = -\alpha D / s_g$, where
\begin{equation}
  s_g
  = \sqrt{
      \frac{\operatorname{tr}(D^\top C_g D)}{\|D\|_F^2}
      + \epsilon_{\mathrm{scale}}
    }.
  \label{eq:app_sg}
\end{equation}

\textbf{Step 1: apply the trace identity.}
By~\eqref{eq:app_trace_id},
\begin{equation}
  \frac{1}{n}\bigl\|X_g\,\Delta W_{\mathrm{Graph}}\bigr\|_F^2
  = \frac{\alpha^2}{s_g^2}\,\operatorname{tr}(D^\top C_g D).
  \label{eq:app_step1}
\end{equation}

\textbf{Step 2: bound $1/s_g^2$.}
Since $\epsilon_{\mathrm{scale}} \ge 0$, the definition~\eqref{eq:app_sg}
gives
\begin{equation}
  s_g^2
  = \frac{\operatorname{tr}(D^\top C_g D)}{\|D\|_F^2} + \epsilon_{\mathrm{scale}}
  \;\ge\; \frac{\operatorname{tr}(D^\top C_g D)}{\|D\|_F^2},
\end{equation}
and therefore
\begin{equation}
  \frac{1}{s_g^2}
  \;\le\; \frac{\|D\|_F^2}{\operatorname{tr}(D^\top C_g D)}.
  \label{eq:app_sg_inv}
\end{equation}

\textbf{Step 3: combine.}
Substituting~\eqref{eq:app_sg_inv} into~\eqref{eq:app_step1},
\begin{align}
  \frac{1}{n}\bigl\|X_g\,\Delta W_{\mathrm{Graph}}\bigr\|_F^2
  &\;\le\;
  \frac{\alpha^2 \|D\|_F^2}{\operatorname{tr}(D^\top C_g D)}
  \cdot \operatorname{tr}(D^\top C_g D)
  = \alpha^2\,\|D\|_F^2.
\end{align}
Taking square roots establishes the inequality
$\frac{1}{\sqrt{n}}\|X_g\,\Delta W_{\mathrm{Graph}}\|_F \le \alpha\|D\|_F$.

\textbf{Equality case.}
When $\epsilon_{\mathrm{scale}} = 0$, the bound~\eqref{eq:app_sg_inv}
holds with equality, so every inequality above is an equality provided the
common factor $\operatorname{tr}(D^\top C_g D)$ is nonzero.  We have
$\operatorname{tr}(D^\top C_g D) = \frac{1}{n}\|X_g D\|_F^2$, which is
zero if and only if $X_g D = 0$.  Hence, when $\epsilon_{\mathrm{scale}} = 0$
and $X_g D \neq 0$, the inequality holds with equality.
\end{proof}

\subsection{Proof of Proposition~\ref{prop:muon_sharpness}
           (Sharpness of Muon Graph Amplification)}
\label{app:proof_prop1}

\begin{proof}
Recall the eigendecomposition $C_g = V\Lambda V^\top$ with eigenvalues
$\lambda_1 \ge \cdots \ge \lambda_{d_{\mathrm{in}}} \ge 0$ and orthonormal
eigenvectors $v_1,\ldots,v_{d_{\mathrm{in}}}$.

\textbf{Construction of $D^\star$.}
Define
\begin{equation}
  D^\star
  = \bigl[v_1 \mid v_2 \mid \cdots \mid v_{d_{\mathrm{out}}}\bigr]
  \;\in\; \R^{d_{\mathrm{in}} \times d_{\mathrm{out}}}.
\end{equation}
Since the eigenvectors are orthonormal, $D^{\star\top} D^\star = I_{d_{\mathrm{out}}}$,
so $D^\star$ has orthonormal columns.  A matrix with orthonormal columns is
the polar factor of itself, making $D^\star$ a feasible polar-factor direction
under Muon's geometry.

\textbf{Representation energy of $D^\star$.}
Applying the trace identity~\eqref{eq:app_trace_id} and the eigendecomposition,
\begin{align}
  \frac{1}{n}\bigl\|X_g D^\star\bigr\|_F^2
  &= \operatorname{tr}(D^{\star\top} C_g\, D^\star) \notag \\
  &= \operatorname{tr}\!\Bigl(
       [v_1 \cdots v_{d_{\mathrm{out}}}]^\top
       V\Lambda V^\top
       [v_1 \cdots v_{d_{\mathrm{out}}}]
     \Bigr) \notag \\
  &= \sum_{i=1}^{d_{\mathrm{out}}} v_i^\top V\Lambda V^\top v_i
   = \sum_{i=1}^{d_{\mathrm{out}}} \lambda_i.
\end{align}
Taking square roots,
\begin{equation}
  \frac{1}{\sqrt{n}}\bigl\|X_g D^\star\bigr\|_F
  = \sqrt{\sum_{i=1}^{d_{\mathrm{out}}} \lambda_i}.
\end{equation}

\textbf{Comparison with $\sqrt{\lambda_{\max}(C_g)}$.}
Because $\lambda_1 = \lambda_{\max}(C_g)$ and all $\lambda_i \ge 0$,
\begin{equation}
  \sum_{i=1}^{d_{\mathrm{out}}} \lambda_i
  \;\ge\; \lambda_1 = \lambda_{\max}(C_g),
\end{equation}
so $\frac{1}{\sqrt{n}}\|X_g D^\star\|_F \ge \sqrt{\lambda_{\max}(C_g)}$.

\textbf{Case $d_{\mathrm{out}} = 1$.}
When $d_{\mathrm{out}} = 1$, we have $D^\star = v_1$ and
$\frac{1}{\sqrt{n}}\|X_g D^\star\|_F = \sqrt{\lambda_1} = \sqrt{\lambda_{\max}(C_g)}$,
so the amplification is attained exactly.

\textbf{Implication for Muon.}
The vanilla Muon bound in~\eqref{eq:muon_graph_perturbation} reads
$\frac{1}{\sqrt{n}}\|X_g \Delta W_{\mathrm{Muon}}\|_F \le \alpha\sqrt{\lambda_{\max}(C_g)}\|D\|_F$.
The construction above shows that $D^\star$, which is feasible under Muon's
Euclidean polar constraint ($D^{\star\top} D^\star = I$), achieves
$\frac{1}{\sqrt{n}}\|X_g D^\star\|_F = \sqrt{\sum_{i=1}^{d_{\mathrm{out}}}\lambda_i}
\ge \sqrt{\lambda_{\max}(C_g)}$.  Hence, the dependence on $\lambda_{\max}(C_g)$
cannot be removed by Muon's matrix-normalization alone, confirming that the
bound is not a loose artifact.
\end{proof}

Moreover, for any positive diagonal matrix $\Sigma$,
the momentum matrix $M=D^\star\Sigma$ has polar factor
$P(M)=D^\star$, confirming that the construction
corresponds to a feasible exact Muon update.

\subsection{Proof of Theorem~\ref{thm:graph_metric_polar}
           (Graph-Metric Polar Optimality)}
\label{app:proof_thm2}

\begin{proof}
Throughout, let $\tilde{M} = \bar{C}_g^{\tau/2} M$ and
$Q = P(\tilde{M})$ be the polar factor of $\tilde{M}$, so that
$D_{g,\tau} = \bar{C}_g^{-\tau/2} Q$ by definition.  Since
$\bar{C}_g \succ 0$ and $\tilde{M}$ has full column rank (by assumption),
$Q$ is unique and satisfies $Q^\top Q = I_{d_{\mathrm{out}}}$.

\textbf{Part 1: $D_{g,\tau}^\top \bar{C}_g^\tau D_{g,\tau} = I$.}

Direct computation:
\begin{align}
  D_{g,\tau}^\top\,\bar{C}_g^\tau\,D_{g,\tau}
  &= \bigl(\bar{C}_g^{-\tau/2} Q\bigr)^\top
     \bar{C}_g^\tau
     \bigl(\bar{C}_g^{-\tau/2} Q\bigr) \notag\\
  &= Q^\top \bar{C}_g^{-\tau/2}\,\bar{C}_g^\tau\,\bar{C}_g^{-\tau/2}\, Q \notag\\
  &= Q^\top Q \notag\\
  &= I.
\end{align}
Here we used that $\bar{C}_g \succ 0$ so its matrix powers commute and
$\bar{C}_g^{-\tau/2}\bar{C}_g^{\tau}\bar{C}_g^{-\tau/2} = I$.

\textbf{Part 2: $D_{g,\tau}$ solves the graph-metric Procrustes problem.}

We wish to show
\begin{equation}
  D_{g,\tau}
  = \argmax_{\substack{D \in \R^{d_{\mathrm{in}} \times d_{\mathrm{out}}}\\
                       D^\top \bar{C}_g^\tau D = I}}
    \operatorname{tr}\!\bigl(D^\top \bar{C}_g^\tau M\bigr).
  \label{eq:app_procrustes}
\end{equation}

\textit{Change of variables.}
Let $\tilde{D} = \bar{C}_g^{\tau/2} D$.  Since $\bar{C}_g \succ 0$, the map
$D \mapsto \tilde{D}$ is a bijection on $\R^{d_{\mathrm{in}} \times d_{\mathrm{out}}}$.

The constraint $D^\top \bar{C}_g^\tau D = I$ transforms as
\begin{equation}
  \tilde{D}^\top \tilde{D}
  = D^\top \bar{C}_g^{\tau/2}\,\bar{C}_g^{\tau/2}\, D \cdot \text{(by } \bar{C}_g \succ 0\text{)}
  \;= D^\top \bar{C}_g^\tau D = I.
\end{equation}
Hence the constraint becomes $\tilde{D}^\top \tilde{D} = I$, i.e.,
$\tilde{D}$ has orthonormal columns.

The objective transforms as
\begin{equation}
  \operatorname{tr}(D^\top \bar{C}_g^\tau M)
  = \operatorname{tr}(D^\top \bar{C}_g^{\tau/2}\,\bar{C}_g^{\tau/2} M)
  = \operatorname{tr}(\tilde{D}^\top \tilde{M}),
\end{equation}
where $\tilde{M} = \bar{C}_g^{\tau/2} M$.

\textit{Apply Lemma~\ref{lem:procrustes}.}
The reformulated problem is
\begin{equation}
  \max_{\tilde{D}^\top \tilde{D} = I}\; \operatorname{tr}(\tilde{D}^\top \tilde{M}),
\end{equation}
which is exactly the orthogonal Procrustes problem.  By
Lemma~\ref{lem:procrustes}, the unique maximiser is
$\tilde{D}^\star = \polar(\tilde{M}) = Q$.

\textit{Map back.}
Returning to the original variable,
$D^\star = \bar{C}_g^{-\tau/2}\tilde{D}^\star = \bar{C}_g^{-\tau/2}Q = D_{g,\tau}$,
which establishes~\eqref{eq:app_procrustes}.

\textbf{Recovery of vanilla Muon at $\tau = 0$.}
When $\tau = 0$, the covariance factors become $\bar{C}_g^0 = I$, so
$\tilde{M} = M$, and $D_{g,0} = P(M)$, which is the exact polar factor
used by vanilla Muon.  The constraint $D^\top D = I$ is then the
standard Euclidean orthonormality constraint, confirming that Theorem~2
generalizes vanilla Muon's polar update.
\end{proof}

\textbf{Output-space interpretation.}
The graph-metric Procrustes solution also minimizes
the distance to the momentum matrix in the same metric:
\begin{equation}
D_{g,\tau}
=\arg\min_{D^\top\bar C_g^\tau D=I}
\left\|\bar C_g^{\tau/2}(D-M)\right\|_F^2.
\end{equation}
At $\tau=1$, let $c=\operatorname{tr}(C_g)/d_{\mathrm{in}}$.
Since $\bar C_g=C_g/c+\epsilon_{\mathrm{cov}}I$,
the graph-metric orthogonality constraint implies
\begin{equation}
\frac{1}{n}(X_gD_{g,1})^\top(X_gD_{g,1})
=c\left(I-\epsilon_{\mathrm{cov}}
D_{g,1}^\top D_{g,1}\right).
\end{equation}
In the unregularized full-rank case, the realized
update therefore has orthogonal columns of equal
squared RMS magnitude. Graph-metric orthogonalization
thus transfers Muon's singular-value balancing from
parameter space to the module-output space.
While this component balances the realized update
directions, graph-aware step normalization separately
controls their overall representation-level magnitude.

\section{Additional \GraphMuon{} Diagnostics}
\label{app:additional_diagnostics}

This appendix provides additional diagnostics for the graph-induced covariance
used by GraphMuon. These results are not required for the main correctness
claims, but they help interpret when graph-aware optimization is expected to
matter and how the graph-awareness parameter $\tau$ affects the update geometry.

\subsection{Effect of Graph-Awareness Strength}
\label{app:tau_interpolation}

GraphMuon uses the interpolation parameter $\tau$ to control the strength of
graph-aware orthogonalization. When $\tau=0$, the covariance factors vanish and
the update reduces to vanilla Muon. As $\tau$ increases, the update direction
increasingly respects the geometry induced by the graph-propagated covariance.

\begin{figure}[t]
    \centering
    \includegraphics[width=0.68\linewidth]{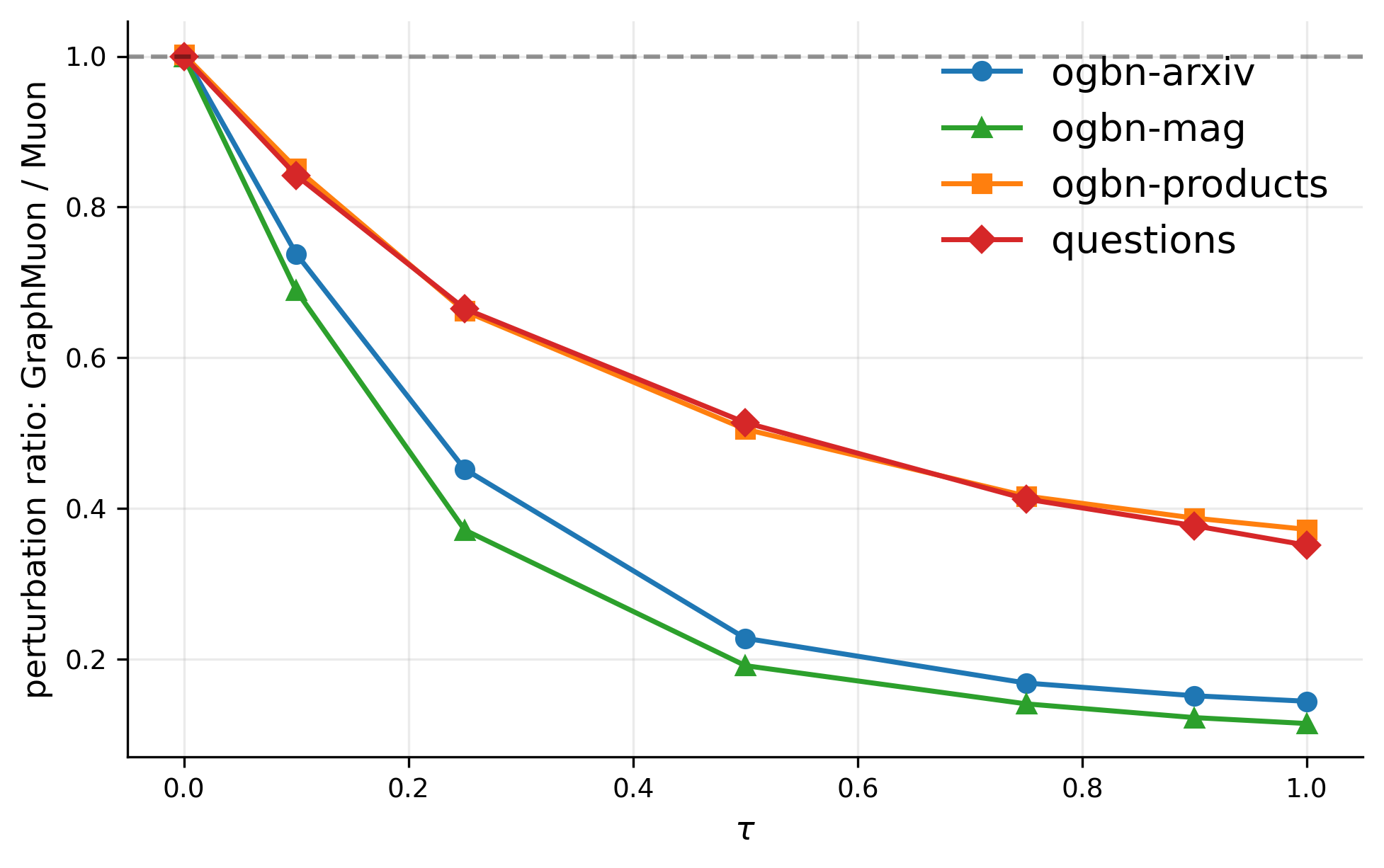}
    \caption{
    Effect of the graph-awareness parameter $\tau$ on representation-level
    perturbations.
    The plot reports the perturbation ratio between GraphMuon and vanilla Muon
    as $\tau$ varies.
    Values below one indicate that the graph-aware direction induces a smaller
    representation-level perturbation than the corresponding Euclidean Muon
    direction.
    Across datasets, increasing $\tau$ generally moves the update away from the
    Euclidean Muon geometry and toward the graph-induced metric, supporting the
    interpretation of $\tau$ as a continuous control over graph awareness.
    }
    \label{fig:tau_interpolation}
\end{figure}

\subsection{Spectral Diagnostics of Graph-Induced Covariance}
\label{app:spectral_diagnostics}

The behavior of GraphMuon depends on the spectral structure of the
graph-induced covariance $C_g=X_g^\top X_g/n$. The dominant eigenvalue measures
the largest graph-amplified direction, while the effective rank measures how
concentrated or distributed the graph-propagated representation is across
spectral directions. These diagnostics help explain both the benefits and the
limitations of graph-aware optimization.

\begin{figure}[t]
    \centering
    \begin{subfigure}{0.49\linewidth}
        \centering
        \includegraphics[width=\linewidth]{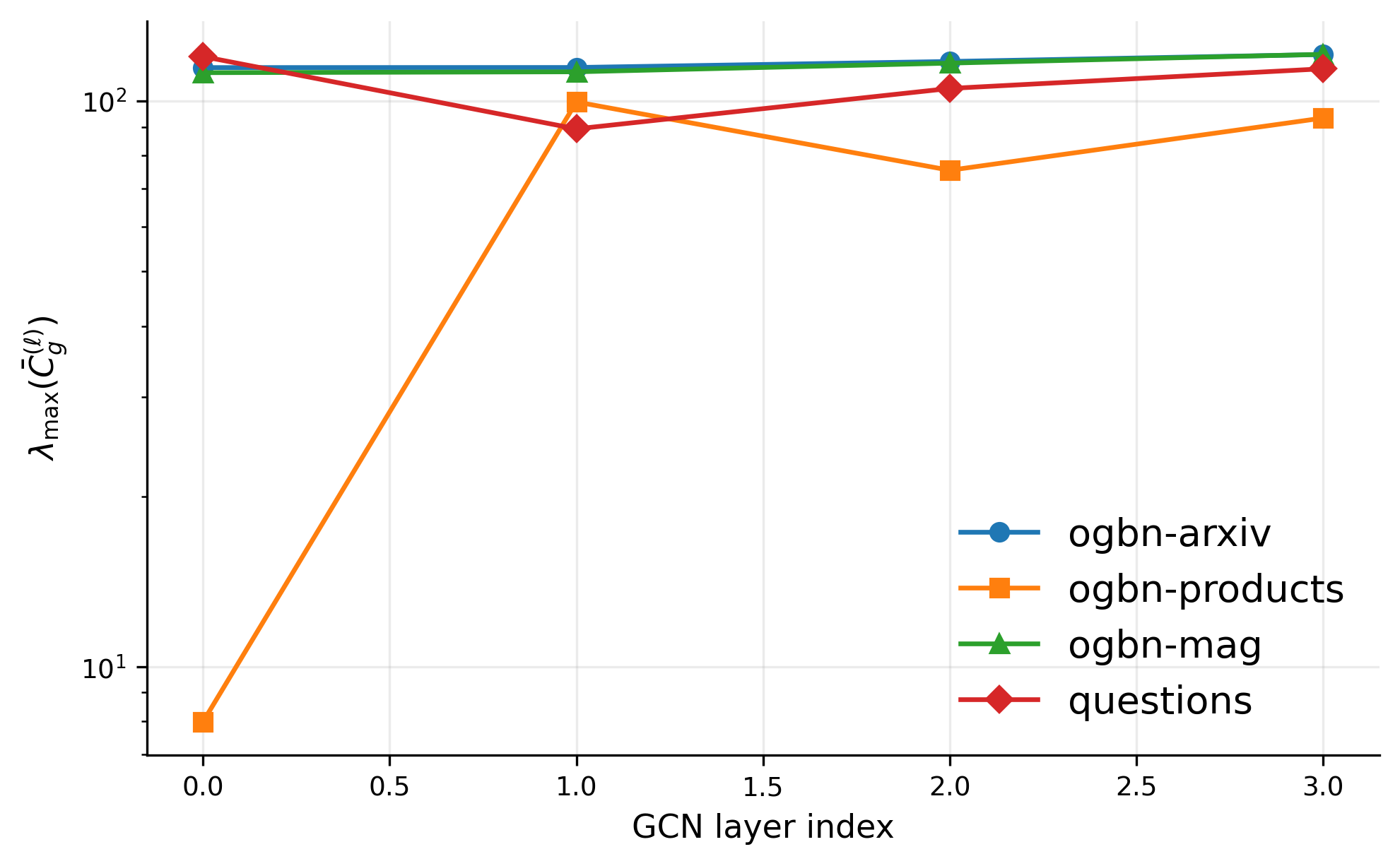}
        \caption{Dominant eigenvalue.}
        \label{fig:lambda_max_per_layer}
    \end{subfigure}
    \hfill
    \begin{subfigure}{0.49\linewidth}
        \centering
        \includegraphics[width=\linewidth]{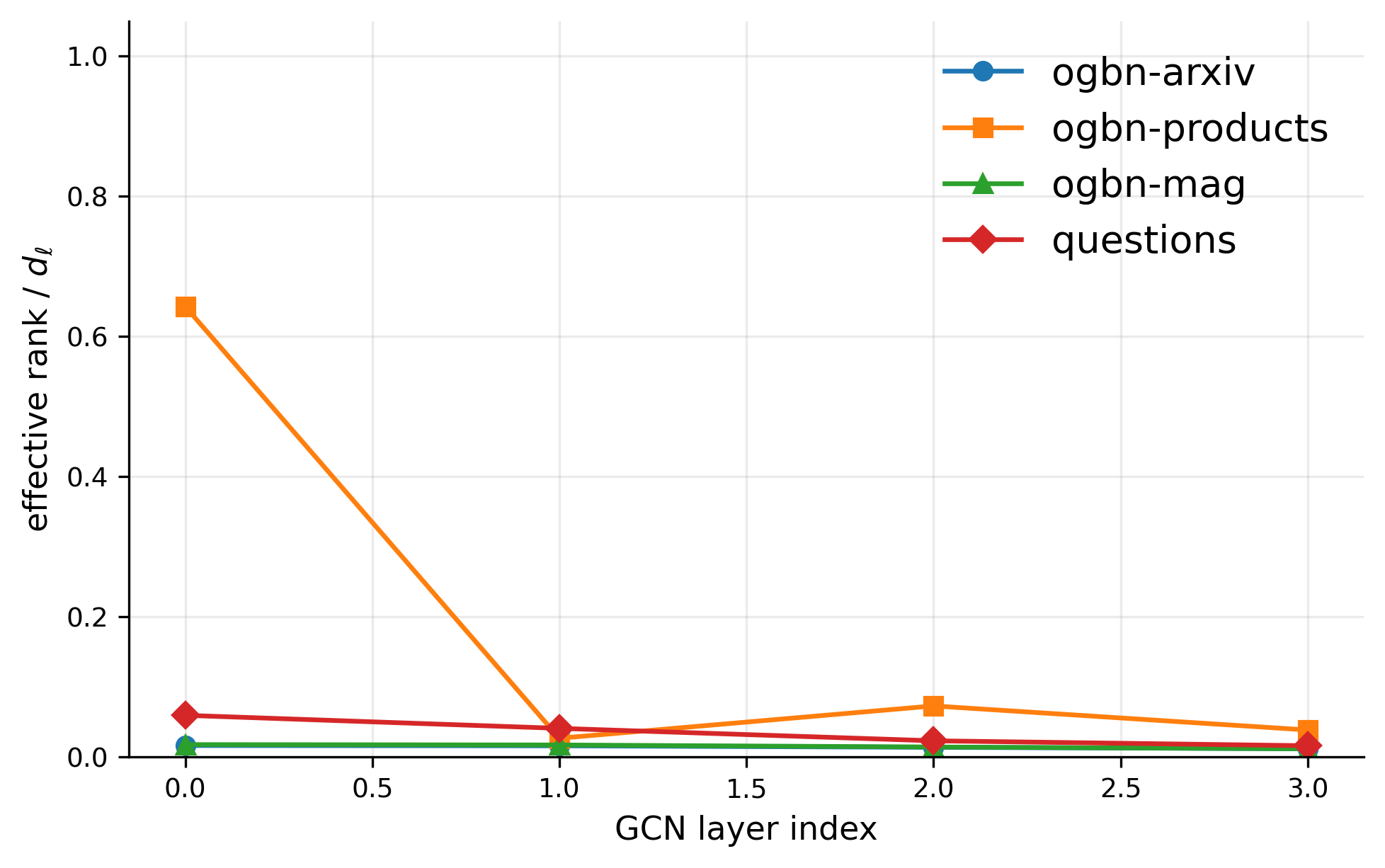}
        \caption{Normalized effective rank.}
        \label{fig:effective_rank_per_layer}
    \end{subfigure}
    \caption{
    Spectral diagnostics of the graph-induced covariance across datasets and
    layers.
    Left: the dominant eigenvalue of $\bar C_g$ identifies high-energy
    graph-propagated directions along which Euclidean-normalized updates can
    induce large representation-level perturbations.
    Right: the normalized effective rank measures how broadly the energy of
    $\bar C_g$ is distributed across directions.
    Lower effective rank indicates that the graph-propagated representation is
    concentrated in fewer dominant modes, making Euclidean matrix geometry less
    aligned with the actual representation geometry.
    Together, these diagnostics support the motivation for GraphMuon: the
    covariance seen by each GNN layer is often anisotropic and layer-dependent,
    so update normalization should account for the graph-propagated input
    geometry.
    }
    \label{fig:spectral_diagnostics}
\end{figure}

\begin{table}[t]
\centering
\caption{Spectral analysis on four-block stochastic block
model (SBM) graphs with varying feature-community alignment.
We report the Rayleigh score of the planted community signal,
its contribution to the graph-induced covariance $C_g$,
and the update energy in dominant $C_g$ directions
before and after GraphMuon scaling.}
\label{tab:sbm_spectral}
\footnotesize
\setlength{\tabcolsep}{4pt}
\begin{tabular}{lccc}
\toprule
\textbf{Alignment} &
\shortstack{\textbf{Community}\\\textbf{Rayleigh}} &
\shortstack{\textbf{Planted share}\\\textbf{of $C_g$}} &
\shortstack{\textbf{Top-$C_g$ energy}\\\textbf{Muon $\rightarrow$ GraphMuon}} \\
\midrule
Strong & 0.708 & 20.3\% & 2.18\% $\rightarrow$ 0.47\% \\
Medium & 0.401 & 9.7\% & 2.37\% $\rightarrow$ 0.43\% \\
Weak   & 0.174 & 2.6\% & 1.06\% $\rightarrow$ 0.18\% \\
\bottomrule
\end{tabular}
\end{table}

\textbf{Controlled spectral analysis.}
To isolate the interaction between graph propagation and
feature alignment, we additionally evaluate GraphMuon on
four-block stochastic block model (SBM) graphs with varying
alignment between node features and planted communities.
Table~\ref{tab:sbm_spectral} shows that as alignment
weakens, the community Rayleigh score decreases from
$0.708$ to $0.174$, while the planted contribution to
$C_g$ decreases from $20.3\%$ to $2.6\%$.
Across all three regimes, GraphMuon substantially reduces
update energy in the dominant graph-amplified directions
relative to Muon. These findings demonstrate that the
graph-induced geometry depends jointly on spectral
propagation strength and feature alignment, and provide
controlled empirical support for graph-aware update
normalization.

\subsection{Sensitivity to the Graph-Metric Exponent \texorpdfstring{$\tau$}{tau}}
\label{app:tau_sensitivity}

We study the sensitivity of GraphMuon to the graph-metric exponent
\(\tau\), which controls the strength of the graph-aware metric
transformation. Smaller values of \(\tau\) keep the update geometry closer
to vanilla Muon, while larger values place more emphasis on the
graph-induced covariance structure. We evaluate several values of
\(\tau\) on representative OGB node-classification benchmarks and plot the
validation accuracy over training.

\begin{figure*}[t]
    \centering
    \begin{subfigure}[t]{0.32\textwidth}
        \centering
        \includegraphics[width=\linewidth]{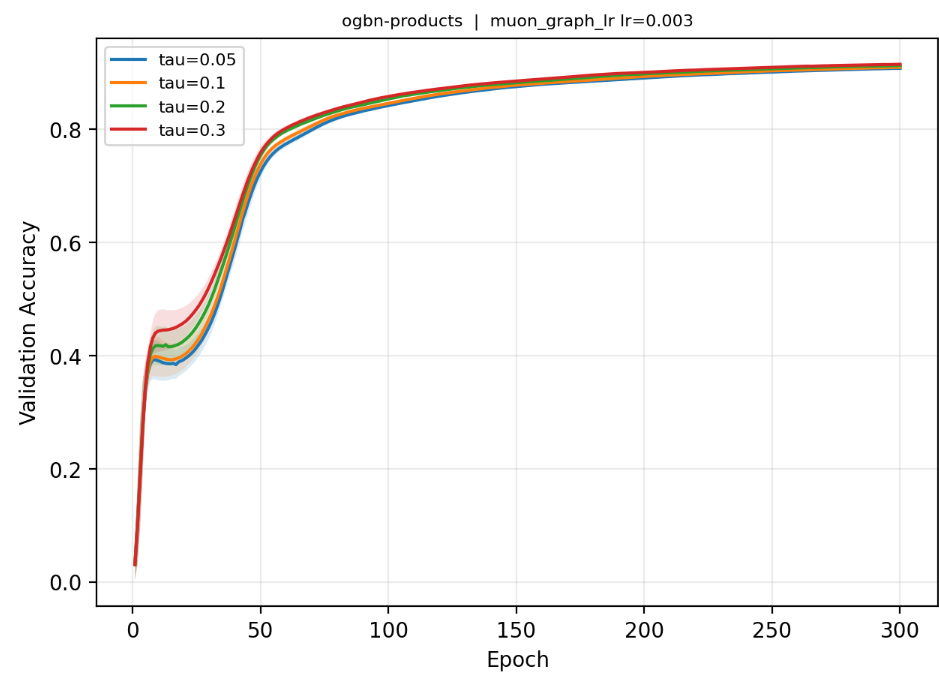}
        \caption{\texttt{ogbn-products}}
        \label{fig:tau_products}
    \end{subfigure}
    \hfill
    \begin{subfigure}[t]{0.32\textwidth}
        \centering
        \includegraphics[width=\linewidth]{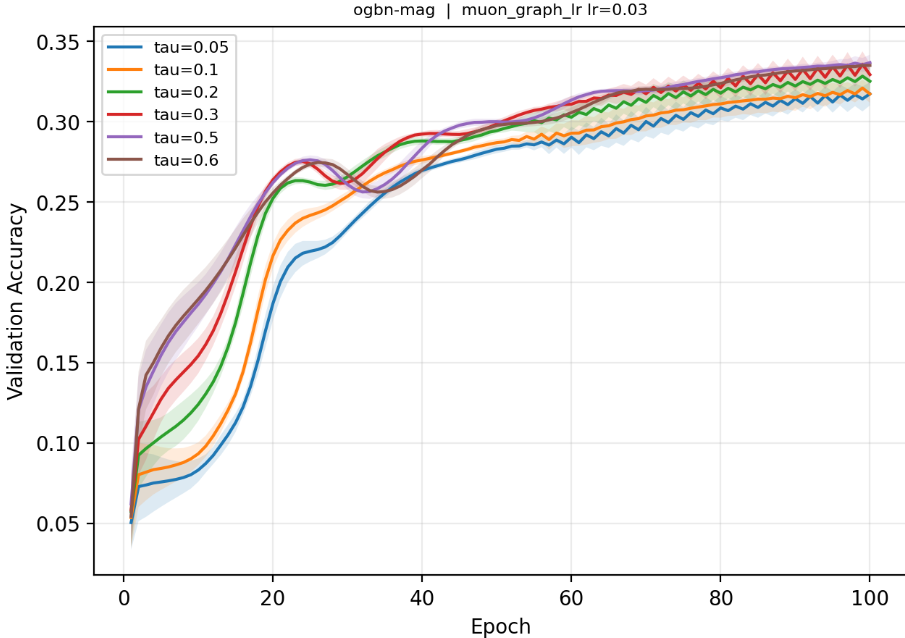}
        \caption{\texttt{ogbn-mag}}
        \label{fig:tau_mag}
    \end{subfigure}
    \hfill
    \begin{subfigure}[t]{0.32\textwidth}
        \centering
        \includegraphics[width=\linewidth]{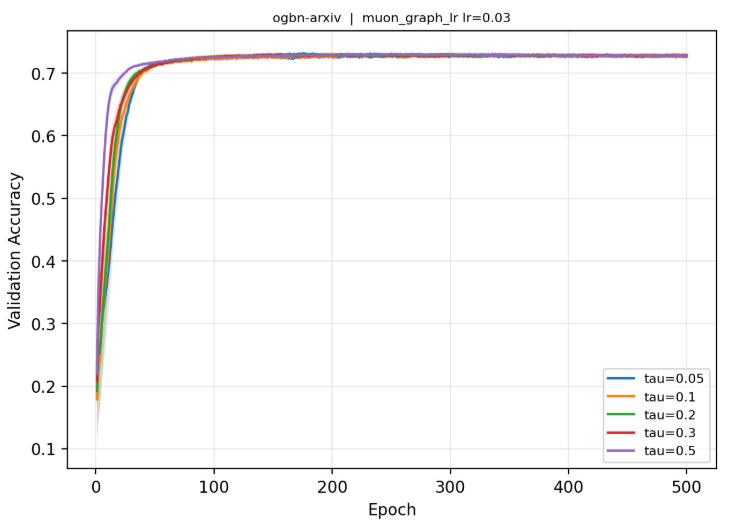}
        \caption{\texttt{ogbn-arxiv}}
        \label{fig:tau_arxiv}
    \end{subfigure}
    \caption{
    Sensitivity of GraphMuon to the graph-metric exponent \(\tau\).
    Each panel shows validation accuracy as a function of training epoch for
    several values of \(\tau\), with the learning rate fixed within each
    dataset. GraphMuon is generally robust to \(\tau\), although the most
    effective value is dataset-dependent. On \texttt{ogbn-arxiv}, all values
    achieve nearly identical final accuracy. On \texttt{ogbn-products},
    moderate values yield slightly faster optimization and marginally better
    final performance. On \texttt{ogbn-mag}, larger values of \(\tau\)
    provide clearer gains, suggesting that stronger graph-aware geometry is
    more beneficial in this setting.
    }
    \label{fig:tau_sensitivity}
\end{figure*}

Figure~\ref{fig:tau_sensitivity} shows that GraphMuon is reasonably robust
to the choice of \(\tau\), although the best value is somewhat
dataset-dependent. On \texttt{ogbn-arxiv}, all tested values produce very
similar final validation accuracy, with only minor differences in early
training speed. On \texttt{ogbn-products}, moderate values
(\(\tau \in [0.1, 0.3]\)) yield slightly stronger early progress and
marginally better final validation performance than the smallest tested
value. On \texttt{ogbn-mag}, the effect of \(\tau\) is more pronounced:
larger values generally improve optimization and lead to better validation
accuracy, while very small values converge more slowly and to slightly
weaker solutions.

Overall, these results indicate that GraphMuon is not overly brittle to the
choice of \(\tau\), but that tuning \(\tau\) can provide additional gains,
especially on more challenging or structurally heterogeneous datasets such
as \texttt{ogbn-mag}. In practice, a small grid over \(\tau\) is sufficient
to identify a strong configuration.

\section{Additional Experimental Setup}
\label{app:experimental_setups}

\subsection{Dataset Statistics}
\label{app:dataset_statistics}

Table~\ref{tab:dataset_statistics} reports the statistics of the
node-classification benchmarks used in our experiments. These datasets cover
homogeneous citation and co-purchase graphs, a heterogeneous academic graph, and
a heterophilous benchmark. For all datasets, we follow the official splits and
evaluation protocols from the corresponding benchmark release.

\begin{table}[h]
\centering
\caption{
Dataset statistics for the node-classification benchmarks. For
\textsc{ogbn-mag}, node and edge counts are reported over all node and relation
types, while features and targets correspond to the target paper nodes.
}
\label{tab:dataset_statistics}
\small
\begin{tabular}{l l r r r r}
\toprule
\textbf{Dataset} 
& \textbf{Graph regime} 
& \textbf{\#Nodes} 
& \textbf{\#Edges} 
& \textbf{\#Features} 
& \textbf{\#Targets} \\
\midrule
\textsc{ogbn-arxiv} 
& Homogeneous citation 
& 169,343 
& 1,166,243 
& 128 
& 40 \\
\textsc{ogbn-products} 
& Homogeneous co-purchase 
& 2,449,029 
& 61,859,140 
& 100 
& 47 \\
\textsc{ogbn-mag} 
& Heterogeneous academic 
& 1,939,743 
& 21,111,007 
& 128 
& 349 \\
\textsc{Questions}
& Heterophilous benchmark
& 48,921
& 153,540
& 301
& 2 \\
\bottomrule
\end{tabular}
\end{table}

\subsection{Architectures}
\label{app:architectures}
Unless stated otherwise, all experiments follow the architecture, data splits,
evaluation protocol, and training setup specified in the original benchmark
papers. This design keeps the benchmark setting fixed and isolates the effect of
the optimizer being compared~\citep{hu2020open, platonov2023critical}.
Table~\ref{tab:backbones} summarizes the backbone used for each dataset.

\begin{table}[t]
\centering
\caption{Dataset--architecture configurations in the
expanded evaluation. The original GCN/GINE settings
are retained, with additional architectures and
graph-level tasks introduced in the rebuttal.}
\label{tab:backbones}
\footnotesize
\setlength{\tabcolsep}{4pt}
\begin{tabular}{lll}
\toprule
\textbf{Dataset} & \textbf{Task} & \textbf{Backbones} \\
\midrule
ogbn-arxiv & Node classification & GCN, GAT, GraphGPS \\
ogbn-mag & Node classification & GCN, GAT, GraphGPS \\
ogbn-products & Node classification & GCN \\
Questions & Node classification & GINE, GAT \\
Cora, Citeseer & Node classification & GCN \\
ogbg-molbace & Graph classification & GINE \\
ogbg-moltox21 & Graph classification & GINE \\
ZINC & Graph regression & GINE, GraphGPS, GraphGCN \\
\bottomrule
\end{tabular}
\end{table}

\textbf{GCN backbone (transductive node classification).}
We use an $L$-layer GCN with symmetric normalized propagation.
Given the adjacency matrix $A$ with added self-loops, we form
$\hat{A} = \tilde{D}^{-1/2}(A + I)\tilde{D}^{-1/2}$
and compute
\begin{equation}
  H^{(l+1)} = \sigma\!\left(\hat{A}\,H^{(l)}\,W^{(l)}\right),
  \qquad l = 0,\ldots,L-2,
\end{equation}
with a linear readout at layer $L$.
Weight matrices are stored in $\mathbb{R}^{d_\mathrm{in} \times d_\mathrm{out}}$
(i.e.\ the propagated feature matrix multiplies on the right)
and initialized with Xavier uniform initialization.
Intermediate layers apply ReLU activation, dropout, and residual
connections whenever $d_\mathrm{in} = d_\mathrm{out}$.
Default settings are depth $L=2$, hidden dimension $d=64$, and
dropout rate $p=0.5$.
Layer normalization is disabled unless noted otherwise.
The final-layer logits are passed to a cross-entropy loss over
the training-mask nodes.

\textbf{GINE backbone (\textsc{questions}).}
For the heterophilous \textsc{questions} benchmark we follow the architecture
of Platonov et al.~\citep{platonov2023critical} and use a GINE
backbone~\citep{hu2020strategies} adapted for transductive node classification.
Each layer applies the update
\begin{equation}
  h_v^{(l+1)} = \mathrm{MLP}^{(l)}\!\left(
    (1 + \varepsilon^{(l)})\,h_v^{(l)}
    + \sum_{u \in \mathcal{N}(v)}
      \mathrm{ReLU}\!\left(h_u^{(l)} + e_{uv}\right)
  \right),
\end{equation}
where $e_{uv}$ are learned edge feature embeddings and $\varepsilon^{(l)}$
is a trainable scalar. Node features are projected to the hidden dimension
by a linear input encoder; edge features are embedded by a learned
\textsc{questions}-specific edge encoder. Unlike the molecular GINE setup,
no \texttt{AtomEncoder}/\texttt{BondEncoder} or graph-level pooling is used:
the final-layer node embeddings are passed directly through a linear head
to produce per-node logits, and a binary cross-entropy loss is computed over
the training-mask nodes following the official \textsc{questions} protocol.
For graph-aware optimizer variants, each GINE layer's first MLP weight matrix
is treated as the graph-conditioned operator, with its graph covariance
computed from the aggregated GINE input
$(1+\varepsilon^{(l)})\,h_v^{(l)} + \sum_{u\in\mathcal{N}(v)}\mathrm{ReLU}(h_u^{(l)}+e_{uv})$;
subsequent MLP weight matrices are treated as standard dense layers and use
the post-activation hidden representation as their input.

\textbf{Heterogeneous graphs.}
For datasets with heterogeneous structure (e.g.\ ogbn-mag) we follow
the standard preprocessing and backbone established for the
corresponding OGB benchmark, including paper-node subgraph
extraction for ogbn-mag.

\subsection{Optimizer Hyperparameters and Implementation Details}
\label{app:hyperparams}

\textbf{Hyperparameter selection protocol.}
All optimizers were tuned using the same validation protocol and the
official validation split for each dataset. For each optimizer and dataset,
the base learning rate was selected from
\[
\{0.001,\;0.003,\;0.01,\;0.03,\;0.1\, ,\;0.15\}.
\]
The configuration with the best validation accuracy was used for test
accuracy reporting. The same selected configuration was then used for
time-to-target evaluation, without additional retuning for the target metric.

\textbf{Best-validation versus fastest-target configurations.}
For optimizer variants reported without a superscript, hyperparameters are
selected by the highest validation accuracy on the official validation split.
Thus, \textsc{GraphMuon} denotes the configuration selected by best validation
performance, using the same model-selection criterion as the other optimizers.

Rows marked with a superscript $*$ report the fastest configuration within
the same hyperparameter grid. Specifically, \textsc{GraphMuon}$^*$ denotes
the GraphMuon configuration that reaches the AdamW best-validation target in
the shortest wall-clock time. This configuration is used only for
time-to-target analysis and is reported separately from the best-validation
configuration to distinguish predictive model selection from optimization
speed selection.

\textbf{Training budget.}
For each dataset, we use the training budget specified by the corresponding
benchmark implementation. Time-to-target is computed only within these
benchmark budgets; runs that do not reach the target within the corresponding
budget are marked as not reaching the target.

\begin{table}[h]
\centering
\caption{Dataset-specific training budgets used in our experiments.}
\label{tab:dataset_budgets}
\begin{tabular}{lcc}
\toprule
Dataset & Benchmark source & Training budget \\
\midrule
\textsc{ogbn-arxiv} & OGB official \citep{hu2020open} & 500 epochs \\
\textsc{ogbn-products} & OGB official \citep{hu2020open} & 300 epochs \\
\textsc{ogbn-mag} & OGB official \citep{hu2020open} & 100 epochs \\
\textsc{questions} & Platonov et al. benchmark \citep{platonov2023critical} & 300 epochs \\
\bottomrule
\end{tabular}
\end{table}

\textbf{Common settings.}
Unless otherwise noted, all experiments share the hyperparameters in
Table~\ref{tab:hparams_common}.

\begin{table}[h]
\centering
\caption{Shared training hyperparameters.}
\label{tab:hparams_common}
\begin{tabular}{lc}
\toprule
Hyperparameter & Value \\
\midrule
Depth $L$ & 2 \\
Hidden dimension $d$ & 64 \\
Dropout $p$ & 0.5 \\
Weight decay $\lambda$ & $5 \times 10^{-4}$ \\
Residual connections & enabled when dimensions match \\
Layer normalization & disabled \\
LR scheduler & none \\
Early-stopping patience & 100 \\
Seeds & 10 \\
\bottomrule
\end{tabular}
\end{table}

\textbf{AdamW.}
AdamW uses $\beta_1 = 0.9$, $\beta_2 = 0.999$,
$\varepsilon = 10^{-8}$, and decoupled weight decay
$\lambda = 5 \times 10^{-4}$.

\textbf{Muon.}
Muon uses momentum $\mu = 0.95$, Nesterov momentum, and
$K=5$ Newton--Schulz orthogonalization iterations. Matrix-valued weight
parameters are updated with Muon. Bias vectors and all non-matrix parameters
are updated with AdamW using the same selected base learning rate and
$\lambda_{\mathrm{other}}=0$.

\textbf{GraphMuon.}
GraphMuon uses the same momentum, Nesterov, Newton--Schulz, and
weight-decay settings as Muon. The graph-awareness parameter is selected
from
\[
\tau \in \{0.1,\;0.2,\;0.3,\;0.5\}
\]
for each dataset using validation accuracy. In our implementation, this
corresponds to the \texttt{muon\_graph\_lr} optimizer variant.

Let $W^{(\ell)} \in \mathbb{R}^{d_{\mathrm{in}}\times d_{\mathrm{out}}}$
denote the weight matrix of layer $\ell$, and let
$G_t^{(\ell)} = \nabla_{W^{(\ell)}}\mathcal{L}_t$ be its stochastic
gradient at step $t$. For readability, we omit the layer superscript below.

\textbf{Nesterov momentum.}
GraphMuon first forms the Muon momentum matrix
\begin{align}
  B_t &= \mu B_{t-1} + G_t, \\
  M_t &= G_t + \mu B_t,
\end{align}
with $\mu=0.95$ and $B_0=0$.

\textbf{Graph covariance.}
For each GNN layer, let
\[
H_{\mathrm{prop}} = X_g = SH
\]
denote the graph-propagated input to the weight matrix. The raw
graph-induced covariance is
\begin{equation}
  C_g = \frac{1}{N}H_{\mathrm{prop}}^\top H_{\mathrm{prop}},
  \label{eq:cov_raw}
\end{equation}
when all $N$ propagated node representations are used. For scalability, when
$N > m$, a subset $H' \in \mathbb{R}^{m\times d}$ is drawn uniformly without
replacement from all $N$ rows of $H_{\mathrm{prop}}$ at every optimizer step,
using a fresh sample for each update. In this case, we estimate
\begin{equation}
  C_g = \frac{1}{m}{H'}^\top H'.
  \label{eq:cov_sampled}
\end{equation}
Both $C_g$ and $\bar{C}$ are computed from this single draw; no second sample
is taken for the step-normalization term.

For graph-metric orthogonalization, we use the trace-normalized and
regularized covariance
\begin{equation}
  \bar{C}
  =
  \frac{C_g}{\mathrm{tr}(C_g)/d_{\mathrm{in}}}
  +
  \varepsilon_{\mathrm{metric}} I.
  \label{eq:cov_normalized}
\end{equation}
A single regularization constant
\[
\varepsilon_{\mathrm{metric}} = 10^{-5}
\]
is used throughout: it is added to the diagonal of $\bar{C}$ for covariance
regularization, used to clamp eigenvalues away from zero when computing
fractional matrix powers, and added inside the graph step norm $s_g$ to
prevent division by zero. All uses share the same value and are denoted
uniformly by $\varepsilon_{\mathrm{metric}}$.

For $d_{\mathrm{in}}\leq 512$, fractional matrix powers of $\bar{C}$ are
computed by eigendecomposition. For larger input dimensions, we use the
diagonal approximation described in Section~\ref{sec:graph_aware_orthogonalization}.

\textbf{Graph-metric orthogonalization.}
GraphMuon maps the momentum matrix into the graph metric, applies
Newton--Schulz orthogonalization, and maps the result back:
\begin{equation}
  O_t
  =
  \bar{C}^{-\tau/2}
  \mathrm{NSK}_{K}
  \left(
    \bar{C}^{\tau/2} M_t
  \right),
  \label{eq:graphmuon_orthog}
\end{equation}
where $K=5$ and $\mathrm{NSK}_{K}$ denotes $K$ Newton--Schulz
orthogonalization steps. Setting $\tau=0$ recovers the vanilla Muon
orthogonalization direction.

\textbf{Graph-aware learning-rate scaling.}
After orthogonalization, GraphMuon rescales the update according to its
induced perturbation under the raw graph covariance:
\begin{equation}
  s_g
  =
  \sqrt{
    \frac{\mathrm{tr}(O_t^\top C_g O_t)}
         {\|O_t\|_F^2}
    +
    \varepsilon_{\mathrm{metric}}
  }.
  \label{eq:graph_lr}
\end{equation}
Here $C_g$ is the raw, unnormalized covariance from
Eq.~\eqref{eq:cov_raw} or Eq.~\eqref{eq:cov_sampled}, not the
trace-normalized covariance $\bar{C}$. Thus, $s_g$ measures the
root-mean-square graph-metric energy of the update direction. Directions
aligned with high-variance graph-propagated features receive a smaller
effective step, while directions aligned with lower-energy graph directions
receive a relatively larger step.

\textbf{Shape-dependent learning-rate correction.}
Following Muon, we apply the rectangular-matrix correction
\begin{equation}
  r(d_{\mathrm{in}}, d_{\mathrm{out}})
  =
  \sqrt{
    \max\left(1,\frac{d_{\mathrm{in}}}{d_{\mathrm{out}}}\right)
  },
  \label{eq:shape_correction}
\end{equation}
which compensates for the RMS scale of semi-orthogonal updates.

\textbf{Parameter update.}
Weight decay is applied in decoupled form before the gradient step, using
the shape-corrected rate and without graph learning-rate scaling:
\begin{equation}
  W_{t+1}
  =
  \bigl(1-\eta\,r\,\lambda\bigr)W_t
  -
  \eta\,r\cdot\frac{O_t}{s_g},
  \label{eq:graphmuon_update_appendix}
\end{equation}
where $r = r(d_{\mathrm{in}},d_{\mathrm{out}})$,
$\lambda = 5\times10^{-4}$ is weight decay, and $\eta$ is the
grid-searched base learning rate. Weight decay is applied only to matrix
weights; bias vectors and other non-matrix parameters are updated by AdamW
with $\lambda_{\mathrm{other}}=0$.

\textbf{Shampoo~\citep{gupta2018shampoo}.}
Shampoo uses momentum $0.0$, $\varepsilon = 10^{-4}$, and preconditioner
update frequency of one step.

\textbf{Natural Gradient~\citep{izadi2020optimization}.}
The natural-gradient baseline uses Tikhonov regularization
$\varepsilon_{\mathrm{kfac}} = 0.01$, covariance update frequency of
50 steps, no running-average smoothing $(\alpha = 1.0)$, and
$\pi$-correction disabled.

\subsection{Efficiency of the Graph-Aware Muon Implementation}
\label{app:efficiency}

We describe the implementation mechanisms used to keep the additional cost of
GraphMuon modest relative to vanilla Muon. The key design choice is that the
optimizer reuses graph-propagated activations already computed during the
ordinary forward pass, and only estimates low-dimensional covariance statistics
from them.

\textbf{Activation reuse.}
For a GCN layer, the forward pass computes the graph-propagated input
\[
    H_{\mathrm{prop}} = \hat{A}H
\]
before applying the learnable weight matrix. Our implementation stores a
detached reference to this tensor as \texttt{cached\_h\_prop}. The optimizer
then reads this cached tensor at step time. Thus, GraphMuon does not perform an
additional forward pass and does not apply an additional graph propagation inside
the optimizer. The extra runtime cost at this stage is only the detach/reference
operation, while the extra memory cost is the lifetime of the cached
$H_{\mathrm{prop}}$ tensor until the optimizer step.

\textbf{Subsampled covariance estimation.}
Let $N$ denote the number of available node representations and let
$m=\texttt{metric\_sample\_size}$ denote the covariance-estimation budget
(default $m=2048$). When $N>m$, we sample $m$ rows of $H_{\mathrm{prop}}$
without replacement and form the covariance estimate
\[
    C_g = \frac{1}{m} H'^{\top}H',
\]
where $H'\in\mathbb{R}^{m\times d_{\mathrm{in}}}$ contains the sampled rows.
This reduces the covariance cost from
$O(Nd_{\mathrm{in}}^2)$ to $O(md_{\mathrm{in}}^2)$. When $N\le m$, the full
set of nodes is used. The same sampled rows are used both for $C_g$ and for the
normalized metric $\bar{C}$, so only one sample is drawn per optimizer step.

\begin{table}[t]
\centering
\caption{Efficiency-related GraphMuon hyperparameters.}
\label{tab:efficiency_hparams}
\begin{tabular}{lcp{7cm}}
\toprule
Hyperparameter & Default & Role \\
\midrule
\texttt{metric\_sample\_size} & 2\,048
  & Maximum number of nodes used to estimate $C_g$ per step.
    Setting it to $0$ disables subsampling and uses all nodes. \\
\texttt{max\_full\_dim} & 512
  & Input dimension threshold for forming the full
    $d_{\mathrm{in}}\times d_{\mathrm{in}}$ covariance. Above this threshold,
    a diagonal approximation is used. \\
\texttt{eigen\_refresh\_steps} & 10
  & Maximum number of optimizer steps for which a cached eigendecomposition of
    $\bar{C}$ may be reused. Setting it to $\le 0$ disables caching. \\
\texttt{eigen\_change\_tol} & 0.05
  & Relative Frobenius-change threshold used to trigger an early refresh of the
    eigendecomposition. \\
\texttt{torch\_foreach} & \texttt{true}
  & Enables PyTorch foreach kernels for weight decay and parameter updates,
    reducing Python and kernel-launch overhead. \\
\bottomrule
\end{tabular}
\end{table}

\textbf{Full and diagonal covariance paths.}
When $d_{\mathrm{in}}\le \texttt{max\_full\_dim}$, GraphMuon forms the full
$d_{\mathrm{in}}\times d_{\mathrm{in}}$ covariance and computes its
eigendecomposition. For larger input dimensions, the implementation falls back
to a diagonal metric. In this case, writing
$c=\mathrm{diag}(H'^{\top}H')/m$, we use the diagonal approximation
\[
    \bar{C}_{\mathrm{diag}}
    =
    \mathrm{diag}\!\left(
    \frac{c}{\overline{c}+\varepsilon_{\mathrm{metric}}}
    + \varepsilon_{\mathrm{metric}}
    \right),
\]
where $\overline{c}$ is the mean diagonal entry. This reduces memory from
$O(d_{\mathrm{in}}^2)$ to $O(d_{\mathrm{in}})$ and avoids the
$O(d_{\mathrm{in}}^3)$ eigendecomposition. Fractional metric powers and
graph-aware step normalization are then applied element-wise. In our standard
benchmark configurations, $d_{\mathrm{in}}\le 128$, so the full covariance path
is used.

\textbf{Eigendecomposition caching.}
For the full covariance path, computing the eigendecomposition of $\bar{C}$
costs $O(d_{\mathrm{in}}^3)$ per layer. GraphMuon caches the eigenvectors and
eigenvalues of $\bar{C}$ in the optimizer state and reuses them when the metric
has not changed substantially. Specifically, if the cached decomposition is less
than \texttt{eigen\_refresh\_steps} steps old and
\[
    \frac{\|\bar{C}_t-\bar{C}_{t-k}\|_F}
         {\|\bar{C}_{t-k}\|_F+\varepsilon_{\mathrm{metric}}}
    \le
    \texttt{eigen\_change\_tol},
\]
the cached decomposition is reused. Otherwise, the eigendecomposition is
recomputed and the cache age is reset. The staleness check costs
$O(d_{\mathrm{in}}^2)$, which is dominated by covariance formation whenever
$m$ is nontrivial. We also cache the power vectors
$\lambda_i^{\tau/2}$ and $\lambda_i^{-\tau/2}$ for the current value of
$\tau$; these are invalidated whenever the eigendecomposition is refreshed.

\textbf{Foreach update path.}
With \texttt{torch\_foreach=true}, weight decay and parameter updates are
applied using PyTorch multi-tensor kernels. Parameters are grouped by their
shape-corrected learning rate $\eta\,r(d_{\mathrm{in}},d_{\mathrm{out}})$, and
all tensors in the same group are updated in a single foreach call. This does
not fuse the graph-aware metric computation itself, but it reduces Python
overhead and GPU kernel-launch latency in the final update stage. 

\textbf{Per-step cost.}
Table~\ref{tab:step_cost} summarizes the per-layer optimizer cost for the
full-covariance path with active eigen caching. The Newton--Schulz
orthogonalization has the same asymptotic cost as in vanilla Muon. The extra
GraphMuon-specific terms are covariance estimation, occasional
eigendecomposition, metric application, and graph-aware step normalization. These
terms are controlled by node subsampling, the full/diagonal covariance switch,
and eigendecomposition caching.

\begin{table}[t]
\centering
\caption{Asymptotic per-layer optimizer cost for the full-covariance GraphMuon
path. Here $m$ is the number of sampled nodes, $d=d_{\mathrm{in}}$, and
$k$ is the number of Newton--Schulz iterations.}
\label{tab:step_cost}
\begin{tabular}{lll}
\toprule
Operation & Cost & Notes \\
\midrule
Read cached $H_{\mathrm{prop}}$
  & $O(1)$
  & Tensor already computed in the forward pass \\
Node subsampling
  & $O(N)$
  & Due to \texttt{torch.randperm} \\
Covariance $C_g$
  & $O(md^2)$
  & Gram matrix of sampled propagated features \\
Staleness check
  & $O(d^2)$
  & Relative Frobenius change of $\bar{C}$ \\
Eigendecomposition
  & $O(d^3)$
  & Amortized by caching \\
Power scales $\lambda^{\pm \tau/2}$
  & $O(d)$
  & Cached for fixed $\tau$ \\
Metric application
  & $O(d^2 d_{\mathrm{out}})$
  & Two applications of $Q(s\odot Q^\top M)$ \\
Newton--Schulz
  & $O(k d d_{\mathrm{out}}\min(d,d_{\mathrm{out}}))$
  & Same order as vanilla Muon \\
  Step normalization
  & $O(d^2 d_{\mathrm{out}})$
  & Computes $\operatorname{tr}(O^\top C_g O)$ \\
Parameter update
  & $O(d d_{\mathrm{out}})$
  & Foreach-batched when enabled \\
\bottomrule
\end{tabular}
\end{table}

\begin{table}[t]
\centering
\caption{Scalability with covariance sampling budget on
\textsc{ogbn-arxiv}. We vary the number of sampled rows
$m$ used to estimate the graph-induced covariance.
Direction cosine and graph-energy error are measured
relative to the full-covariance reference.}
\label{tab:scalability_sampling}
\footnotesize
\setlength{\tabcolsep}{4pt}
\begin{tabular}{rccccc}
\toprule
\textbf{Rows $m$} &
\shortstack{\textbf{Best val.}\\\textbf{loss} $\downarrow$} &
\shortstack{\textbf{Epochs to}\\\textbf{target} $\downarrow$} &
\shortstack{\textbf{Test}\\\textbf{acc.} $\uparrow$} &
\shortstack{\textbf{Direction}\\\textbf{cosine} $\uparrow$} &
\shortstack{\textbf{Graph-energy}\\\textbf{error} $\downarrow$} \\
\midrule
512   & 0.8591 & 153 & 0.7187 & 0.9996 & 0.0098 \\
1,024 & 0.8599 & 151 & \textbf{0.7203} & 0.9998 & 0.0045 \\
2,048 & 0.8592 & 148 & 0.7196 & 0.9999 & 0.0024 \\
4,096 & 0.8598 & \textbf{145} & 0.7198 & \textbf{1.0000} & \textbf{0.0012} \\
\bottomrule
\end{tabular}
\end{table}

\textbf{Scalability with covariance sampling.}
GraphMuon reuses the propagated representations computed
during the forward pass, avoiding additional graph propagation.
For a covariance sampling budget $m$ and input dimension $d$,
covariance construction requires $O(md^2)$ operations,
while cached eigendecomposition adds an amortized
$O(d^3/r)$ cost for a refresh interval of $r$ steps.
For dimensions above the full-metric threshold, the
diagonal approximation avoids the cubic eigendecomposition
cost.

Table~\ref{tab:scalability_sampling} evaluates the effect of
increasing $m$ eightfold, from 512 to 4,096 rows.
Optimization remains stable, reaching the fixed validation-loss
target within 145--153 epochs across all budgets.
Best validation loss varies only between 0.8591 and 0.8599,
while test accuracy remains between 0.7187 and 0.7203.
Larger sampling budgets improve approximation fidelity:
direction cosine increases from 0.9996 to 1.0000, while
graph-energy error decreases from 0.0098 to 0.0012.
These results indicate that accurate graph-aware updates
can be obtained using relatively small covariance samples.

The large-scale memory profile in Table~\ref{tab:memory_footprint}
provides complementary evidence on \textsc{ogbn-products}.
GraphMuon adds approximately 0.005 GB of optimizer-owned
memory relative to Muon, without increasing peak CUDA memory,
and incurs only 2.8\% additional per-step runtime
(1.411 versus 1.372 seconds).
Thus, covariance sampling and eigendecomposition caching
keep the additional graph-aware optimization overhead modest
in the evaluated settings.

\subsection{Hardware, Software, and Baseline Implementations}
\label{app:hardware_implementation}

All experiments were run on a single NVIDIA RTX 6000 GPU. GraphMuon was
implemented in PyTorch and integrated into a shared training pipeline used by all
optimizers. Unless stated otherwise, all methods use the same model
architectures, dataset splits, input features, batching scheme, evaluation
protocol, and hyperparameter tuning budget.

For baselines, AdamW is the standard PyTorch implementation
(\texttt{torch.optim.AdamW}). Muon and GraphMuon are implemented directly in
PyTorch within our training code. Shampoo is implemented using the
\texttt{pytorch-optimizer} library. For any remaining optimizer baselines, we
use the official or original reference implementations when available, adapting
only the optimizer interface required for compatibility with our training code.
These adaptations do not change the mathematical update rules of the baselines.

Runtime is measured on the same hardware for all methods. In time-to-target
experiments, wall-clock time is measured until the validation target is first
reached. One-time dataset preprocessing is excluded from the timing. We
synchronize CUDA before recording timed intervals to avoid artifacts from
asynchronous GPU execution. All reported averages and standard deviations are
computed over the same set of random seeds.

\textbf{Software environment.}
All experiments were conducted with up-to-date stable versions of Python,
PyTorch \citep{paszke2019pytorch} (offered under BSD-3 Clause license), CUDA, PyTorch Geometric \citep{fey2019fast} (offered under MIT license), OGB, and \texttt{pytorch-optimizer} as of May
2026.

\subsection{Memory Footprint}
\label{app:memory_footprint}

We profile the memory footprint of GraphMuon relative to AdamW and vanilla Muon
on an \texttt{ogbn-products}-scale input configuration with approximately
\(1.2\)M nodes and hidden dimension \(512\). Table~\ref{tab:memory_footprint}
reports optimizer state memory, GraphMuon-specific graph-cache memory,
optimizer-owned memory, persistent CUDA memory, peak CUDA memory, and average
step time. Optimizer-owned memory includes both optimizer state and any
GraphMuon-specific cached metric quantities. Persistent CUDA memory measures the
memory that remains allocated during training, while peak CUDA memory reports
the maximum CUDA allocation observed during the profiled step.

\begin{table}[t]
\small
\centering
\caption{
Memory footprint and per-step runtime on an \texttt{ogbn-products}-scale input
configuration with approximately \(1.2\)M nodes and hidden dimension \(512\).
GraphMuon increases optimizer-owned memory because it stores a small
graph-metric cache, but the absolute overhead is only a few megabytes.
Persistent CUDA memory is nearly unchanged, and peak CUDA memory is identical
across the profiled optimizers.
}
\label{tab:memory_footprint}
\begin{tabular}{lcccccc}
\toprule
Optimizer
& State 
& Graph cache 
& Owned 
& Persistent 
& Peak CUDA 
& Step time (s) \\
\midrule
AdamW
& 0.0025
& 0.000
& 0.0025
& 15.807
& 28.211
& 1.076 \\
Muon
& 0.0013
& 0.000
& 0.0013
& 15.806
& 28.211
& 1.372 \\
GraphMuon
& 0.0013
& 0.005
& 0.0063
& 15.811
& 28.211
& 1.411 \\
\bottomrule
\end{tabular}
\end{table}

GraphMuon adds \(0.0050\) GB of graph-cache memory in this profiling run,
bringing its total optimizer-owned memory to \(0.0063\) GB. Although this is
larger than AdamW in relative terms, the absolute increase is small compared to
the total training memory. Persistent CUDA memory increases only from
\(15.8077\) GB for AdamW to \(15.8115\) GB for GraphMuon, corresponding to a
negligible end-to-end memory increase. Peak CUDA memory is identical across all
three optimizers at \(28.2111\) GB.

GraphMuon has a slightly higher per-step time than vanilla Muon in this setting
(\(1.411\)s vs.\ \(1.372\)s), reflecting the additional covariance estimation,
metric application, and graph-aware normalization. This overhead is small in
absolute terms and does not materially increase peak GPU memory usage.

\subsection{Selected GraphMuon Configurations}
\label{app:selected_graphmuon_configs}

Table~\ref{tab:selected_graphmuon_configs} reports the learning rate and
graph-awareness strength $\tau$ used for the selected GraphMuon configurations.
For each dataset, GraphMuon denotes the configuration selected by best validation
accuracy, while GraphMuon$^\ast$ denotes the fastest configuration that reaches
AdamW's validation target. If no GraphMuon$^\ast$ row is shown, the two
selection criteria choose the same configuration.

\begin{table}[h]
\centering
\caption{
Selected learning rate and graph-awareness strength $\tau$ for GraphMuon.
GraphMuon denotes the validation-selected configuration, while GraphMuon$^\ast$
denotes the fastest configuration reaching AdamW's validation target.
}
\label{tab:selected_graphmuon_configs}
\begin{tabular}{llcc}
\toprule
Dataset & Optimizer & Learning rate & $\tau$ \\
\midrule
ogbn-arxiv
& GraphMuon
& 0.01 & 0.1 \\

ogbn-mag
& GraphMuon
& 0.10 & 0.5 \\
& GraphMuon$^\ast$
& 0.15 & 0.5 \\

ogbn-products
& GraphMuon
& 0.01 & 0.1 \\
& GraphMuon$^\ast$
& 0.03 & 0.3 \\

questions
& GraphMuon
& 0.03 & 0.5 \\
& GraphMuon$^\ast$
& 0.03 & 0.3 \\
\bottomrule
\end{tabular}
\end{table}

\subsection{Additional Architecture and Task Results}
\label{app:architecture_results}

We extend the time-to-target evaluation across 15 configurations,
covering GCN, GAT, GINE, GraphGPS, and GraphGCN, as well as
node classification, graph classification, and graph regression.
For each configuration, the target is the minimum mean
validation loss attained by AdamW. We report the first epoch
and corresponding cumulative wall-clock time at which each
optimizer reaches this target.

\begin{table}[t]
\centering
\caption{Complete validation-loss time-to-target results
across 15 dataset--architecture configurations.
Each optimizer entry reports epochs / wall-clock seconds.
Speedups are measured against AdamW's wall-clock time;
\textbf{bold} indicates configurations where GraphMuon
achieves the lowest time-to-target. -- denotes failure
to reach the target within the training budget.}
\label{tab:architecture_full}
\scriptsize
\setlength{\tabcolsep}{3pt}
\resizebox{\textwidth}{!}{%
\begin{tabular}{llrcrcrr}
\toprule
\textbf{Dataset} &
\textbf{Arch.} &
\textbf{Target} &
\textbf{AdamW (ep/s)} &
\textbf{Muon (ep/s)} &
\textbf{Muon $\times$} &
\textbf{GraphMuon (ep/s)} &
\textbf{GraphMuon $\times$} \\
\midrule
ogbn-arxiv & GAT &
1.0118 & 122 / 1478.8 & -- & -- &
\textbf{12 / 161.2} & \textbf{9.17} \\

ogbn-arxiv & GCN &
0.8753 & 433 / 30.3 & 188 / 16.6 & 1.83 &
\textbf{164 / 13.1} & \textbf{2.32} \\

ogbn-arxiv & GraphGPS &
1.7066 & 24 / 1345.2 & 2 / 121.1 & 11.11 &
\textbf{1 / 83.7} & \textbf{16.07} \\

ogbn-mag & GAT &
2.7167 & 69 / 1857.4 & 6 / 160.4 & 11.58 &
\textbf{3 / 110.2} & \textbf{16.85} \\

ogbn-mag & GCN &
2.6930 & 100 / 58.3 & 65 / 38.1 & 1.53 &
\textbf{41 / 24.1} & \textbf{2.42} \\

ogbn-mag & GraphGPS &
3.0715 & 21 / 1978.3 & 3 / 406.3 & 4.87 &
\textbf{1 / 223.9} & \textbf{8.84} \\

ogbn-products & GCN &
0.3182 & 300 / 1663.5 & -- & -- &
\textbf{226 / 1257.3} & \textbf{1.32} \\

cora & GCN &
0.6873 & 18 / 0.0922 & 313 / 1.8 & 0.05 &
\textbf{10 / 0.0807} & \textbf{1.14} \\

questions & GAT &
0.1217 & 481 / 15.2 & 320 / 10.2 & 1.49 &
\textbf{167 / 7.1} & \textbf{2.14} \\

ogbg-molbace & GINE &
0.6212 & 65 / 59.1 & -- & -- &
\textbf{22 / 29.8} & \textbf{1.98} \\

ogbg-moltox21 & GINE &
0.2946 & 9 / 22.3 & 6 / 22.6 & 0.99 &
\textbf{4 / 18.6} & \textbf{1.20} \\

ZINC & GINE &
0.2502 & 293 / 946.6 & 242 / 1038.4 & 0.91 &
\textbf{106 / 679.0} & \textbf{1.39} \\

ZINC & GraphGPS &
0.1584 & 179 / 3176.7 & 78 / 2062.1 & 1.54 &
\textbf{34 / 1192.1} & \textbf{2.66} \\

citeseer & GCN &
1.0150 & 14 / 0.0738 & 24 / 0.2 & 0.37 &
11 / 0.0904 & 0.82 \\

ZINC & GraphGCN &
0.2921 & 490 / 1487.8 & -- & -- &
458 / 1566.1 & 0.95 \\
\midrule
\multicolumn{6}{r}{\textbf{Mean GraphMuon speedup (15 configurations)}} &
& \textbf{4.62} \\
\bottomrule
\end{tabular}%
}
\end{table}

Across all 15 configurations, GraphMuon reaches AdamW's
validation-loss target in fewer epochs, and achieves the lowest
wall-clock time-to-target in 13 of 15 configurations.
It reaches the target in all 15 configurations, compared with
11 of 15 for Muon. The arithmetic-mean wall-clock speedup
over AdamW is $4.62\times$ across all configurations.
The strongest improvements include GAT on
\textsc{ogbn-mag} ($16.85\times$) and GraphGPS on
\textsc{ogbn-arxiv} ($16.07\times$).

The gains are not universal in wall-clock time. On Citeseer
with GCN and ZINC with GraphGCN, GraphMuon requires fewer
epochs but is slightly slower than AdamW in cumulative time.
These cases illustrate that improved epoch-level optimization
does not always translate into wall-clock acceleration,
particularly when optimizer overhead is substantial relative
to the model's training cost.

\begin{table}[t]
\centering
\scriptsize
\setlength{\tabcolsep}{2.0pt}
\caption{
Time-to-target comparison. The target is AdamW's best validation accuracy on
each dataset. Results are mean $\pm$ standard deviation across seeds. Speedup is
AdamW mean wall-clock time divided by optimizer mean time-to-target. Unmarked
rows are selected by best validation accuracy; $^\ast$ marks the fastest
configuration reaching AdamW's target. If no $^\ast$ appears, the two coincide.
``--'' indicates failure to reach the target within budget.
}
\label{tab:ogb_time_to_target_all}
\resizebox{\linewidth}{!}{
\begin{tabular}{l l c c c c}
\toprule
Dataset & Optimizer & Test Acc. & Epochs to Target & Walltime (s) & Speedup vs. AdamW \\
\midrule

\multirow{7}{*}{ogbn-arxiv}
& AdamW
& $0.7222 \pm 0.0022$
& $433.4 \pm 59.3$
& $30.427 \pm 4.010$
& $1.00\times$ \\
& Muon
& $0.7197 \pm 0.0017$
& $284.5 \pm 26.8$
& $19.734 \pm 2.176$
& $1.54\times$ \\
& Muon$^\ast$
& $0.7200 \pm 0.0010$
& $198.4 \pm 54.3$
& $13.442 \pm 3.677$
& $2.26\times$ \\
& Natural Gradient
& $0.7194 \pm 0.0005$
& $264.5 \pm 15.6$
& $17.563 \pm 1.106$
& $1.73\times$ \\
& Shampoo
& $0.7177 \pm 0.0027$
& -- & -- & -- \\
& \method{} 
& $0.7196 \pm 0.0009$
& $130.2 \pm 8.8$
& $10.171 \pm 0.781$
& $2.99\times$ \\
\midrule

\multirow{7}{*}{ogbn-mag}
& AdamW
& $0.3267 \pm 0.0031$
& $90.1 \pm 9.2$
& $52.576 \pm 5.398$
& $1.00\times$ \\
& Muon
& $0.3441 \pm 0.0081$
& $43.6 \pm 6.8$
& $25.587 \pm 3.977$
& $2.05\times$ \\
& Muon$^\ast$
& $0.3426 \pm 0.0051$
& $42.7 \pm 6.4$
& $25.068 \pm 3.763$
& $2.10\times$ \\
& Natural Gradient
& $0.3457 \pm 0.0019$
& $35.8 \pm 2.0$
& $20.986 \pm 1.172$
& $2.51\times$ \\
& Shampoo
& $0.1632 \pm 0.0108$
& -- & -- & -- \\
& \method{} (Ours) 
& $0.3590 \pm 0.0048$
& $25.8 \pm 1.4$
& $15.207 \pm 0.825$
& $3.46\times$ \\
& \method{}$^\ast$ (Ours)
& $0.3532 \pm 0.0093$
& $18.6 \pm 2.3$
& $11.067 \pm 1.359$
& $4.75\times$ \\
\midrule

\multirow{7}{*}{ogbn-products}
& AdamW
& $0.7455 \pm 0.0029$
& $266.0 \pm 28.7$
& $1474.944 \pm 159.170$
& $1.00\times$ \\
& Muon
& $0.7462 \pm 0.0031$
& $165.4 \pm 9.0$
& $915.521 \pm 49.218$
& $1.61\times$ \\
& Muon$^\ast$
& $0.7417 \pm 0.0031$
& $148.7 \pm 13.7$
& $824.160 \pm 75.388$
& $1.79\times$ \\
& Natural Gradient
& $0.7462 \pm 0.0027$
& $176.0 \pm 26.4$
& $979.136 \pm 146.911$
& $1.51\times$ \\
& Shampoo
& $0.6004 \pm 0.0131$
& -- & -- & -- \\
& \method{} (Ours)
& $0.7598 \pm 0.0032$
& $209.5 \pm 7.3$
& $1165.575 \pm 40.489$
& $1.27\times$ \\
& \method{}$^\ast$ (Ours)
& $0.7458 \pm 0.0043$
& $124.8 \pm 8.2$
& $694.053 \pm 45.522$
& $2.13\times$ \\
\midrule

\multirow{7}{*}{questions}
& AdamW
& $0.9707 \pm 0.0004$
& $226.0 \pm 135.8$
& $2.834 \pm 1.649$
& $1.00\times$ \\
& Muon
& $0.9708 \pm 0.0002$
& $94.3 \pm 7.7$
& $1.256 \pm 0.106$
& $2.26\times$ \\
& Muon$^\ast$
& $0.9702 \pm 0.0005$
& $55.0 \pm 0.0$
& $0.747 \pm 0.000$
& $3.79\times$ \\
& Natural Gradient
& $0.9707 \pm 0.0003$
& $35.5 \pm 9.2$
& $0.463 \pm 0.122$
& $6.12\times$ \\
& Shampoo
& $0.9709 \pm 0.0001$
& $72.5 \pm 13.4$
& $6.317 \pm 1.260$
& $0.45\times$ \\
& \method{} (Ours)
& $0.9709 \pm 0.0002$
& $14.8 \pm 3.6$
& $0.436 \pm 0.106$
& $6.50\times$ \\
& \method{}$^\ast$ (Ours)
& $0.9709 \pm 0.0002$
& $13.0 \pm 0.8$
& $0.384 \pm 0.025$
& $7.38\times$ \\
\bottomrule
\end{tabular}
}
\end{table}

\newpage

\end{document}